\documentclass[pdflatex,sn-mathphys-num]{sn-jnl}

\usepackage{array}
\usepackage{graphicx}
\usepackage{multirow}
\usepackage{amsmath,amssymb,amsfonts}
\usepackage{amsthm}
\usepackage{mathrsfs}
\usepackage[title]{appendix}
\usepackage{xcolor}
\usepackage{textcomp}
\usepackage{manyfoot}
\usepackage{booktabs}
\usepackage{algorithm}
\usepackage{algorithmicx}
\usepackage{algpseudocode}
\usepackage{listings}
\usepackage{subcaption}

\begin{document}

\title[Article Title]{Assessment of Submillimeter Precision via Structure from Motion Technique in Close-Range Capture Environments}

\author*[1]{\fnm{Francisco Roza de} \sur{Moraes}}\email{franciscormoraes@gmail.com}

\author[1]{\fnm{Irineu} \sur{da Silva}}\email{irineu@usp.br}
\equalcont{These authors contributed equally to this work.}

\affil[1]{\orgdiv{Department of Transportation Engineering (EESC)}, \orgname{University of São Paulo}, \orgaddress{ \city{São Carlos}, \postcode{13566-590}, \state{São Paulo}, \country{Brazil}}}

\abstract{Creating 3D models through the Structure from Motion technique is a recognized, efficient, cost-effective structural monitoring strategy. This technique is applied in several engineering fields, particularly for creating models of large structures from photographs taken a few tens of meters away. However, discussions about its usability and the procedures for conducting laboratory analysis, such as structural tests, are rarely addressed. This study investigates the potential of the SfM method to create submillimeter-quality models for structural tests, with short-distance captures. A series of experiments was carried out, with photographic captures at a 1-meter distance, using different quality settings: camera calibration model, Scale Bars dispersion, overlapping rates, and the use of vertical and oblique images. Employing a calibration model with images taken over a test board and a set of Scale Bars (SB) appropriately distributed over the test area, an overlap rate of 80 percent, and the integration of vertical and oblique images, RMSE values of approximately 0.1 mm were obtained. This result indicates the potential application of the technique for 3D modeling with submillimeter positional quality, as required for structural tests in laboratory environments.}


\maketitle

\section{Introduction}\label{sec1}

The Structure from Motion (SfM) technique is a Computer Vision approach based on photogrammetric principles, designed to extract high-resolution 3D data by combining two-dimensional images of objects or scenes. This method employs low-cost digital cameras and minimizes the need for extensive supervision and technical expertise. SfM holds significant academic value as it generates elaborate and accurate 3D representations for different applications and analyses.

This computational approach, when combined with 3D positioning techniques or control points and digital imaging systems, revolutionized several areas of engineering that involve three-dimensional modeling. For instance, it has been extensively employed in areas such as topographic surveying \cite{anderson2019,carrera-Hernández2020,stott2020}, erosion analysis \cite{smith2015,warrick2017,zimmer2018}, river flow analysis \cite{morgan2017,scaioni2015}, geological studies \cite{nolan2015,wu2020}, archeology \cite{cucchiaro2020,moyano2020}, and various other fields. By offering an accessible and cost-effective solution, SfM has significantly advanced the capabilities of these disciplines.

The SfM technique in Engineering typically aims to acquire 3D data with a precision level of a centimeter or below, which is ideal for large-scale geographic and geomorphological mapping. Nevertheless, there are few studies in the literature addressing the attainment of millimeter or submillimeter precision, with a predominant focus on laboratory analyses such as structural tests and geotechnical assessments. The challenge lies in defining suitable procedures and parameters for SfM, alongside the need for highly accurate positional references to achieve the desired precision.

After analyzing several high-resolution 3D data capture studies, such as those referenced in \cite{nesbit2019,verma2019}, and conducting laboratory experiments at the Structures and Transportation Laboratories of São Carlos School of Engineering, specific quality parameters were established for SfM to achieve submillimeter precision. These parameters include camera calibration, Scale Bar characteristics, overlapping rates, image set combination, and the optimal number of images.

This study aimed to explore the impact of varying these parameters on the positional accuracy of SfM processing. To evaluate the applicability of this technique in modeling structural tests, which require submillimeter precision or less, we conducted a series of photographic captures at one meter from a wooden board.

We opted for a wooden board as a surrogate for a structural specimen due to its widespread use in structural testing and its distinctive surface texture, which aids in identifying Tie Points during the SfM process. Similarly, we selected the capture distance to achieve submillimeter precision and adhere to safety constraints inherent in structural testing protocols.
Our experiments highlighted the necessity of using a pre-calibration model for cameras, utilizing an environment that mirrors the subject under study. Additionally, they emphasized the importance of capturing images vertically and obliquely with an overlap rate of 80\% to accurately represent the region of interest, thereby making this technique applicable for modeling structural tests.
 
\section{Literature Review}\label{sec2}
\subsection{SfM Model Quality Elements}\label{subsec1}

Combining affordable digital imaging systems, such as action cameras and cell phones, with transportation platforms like UAVs has enhanced the effectiveness of acquiring data for engineering applications. This integration, with user-friendly 3D modeling software, has led to the widespread adoption of SfM across various survey fields.
In \cite{pena2019}, the authors utilized the SfM technique, employing images captured by a standard camera, for documenting rock art. Consequently, the 3D models produced demonstrated geometrically significant results compared to the Terrestrial Laser Scanner (TLS) technique, which is known for its quality but also for its high cost.

The SfM technique, like Photogrammetry, uses overlapping image sets for 3D modeling but provides greater automation throughout the process \cite{carrivick2016}. However, with increased automation comes higher susceptibility to sets of images with a high overlapping value, requiring a greater correlation in information between images \cite{westoby2012}.

Several studies \cite{gerke2016,james2014,morgan2017,rupnik2015} explored the impact of varying image overlap on SfM modeling, particularly for UAV-based long-distance photography, enhancing the positional quality of 3D products. In \cite{nesbit2019}, using 70\% and 90\% overlap rates resulted in markedly improved Root Mean Squared Error (RMSE) values compared to standard rates from Photogrammetry (60\%).

While increased overlap improves positional quality, scene geometry and the distribution of SB or Ground Control Points (GCP) also impact product quality. \cite{ridolfi2017} examined combinations of vertical and oblique images, inclined over 30°, resulting in improved positional accuracy and detailed features, particularly for close-range captures.

In \cite{verma2019} and \cite{ridolfi2017}, researchers analyzed the effects of quantity and patterns of SBs and Ground Control Points (GCPs). The initial study examined the use of laboratory-defined sets of GCPs and SBs for rock modeling. In the second study, researchers strategically placed numerous GCPs along a dam body, employing varied reference point patterns and density to enhance quality.

Both studies and others focusing on SfM modeling quality advocate for an increased concentration of GCPs or SBs in regions with rugged geometry or detailed surfaces. They emphasize a wide distribution across the object's surface, varying the elevation (GCPs) and directions (SBs) of reference points to enhance model quality and positional accuracy.

In \cite{ridolfi2017}, the researchers also investigated the impact of changes in the overlapping rate and GCP arrangement, along with experimenting with different camera calibration models, particularly in UAV-based photographic captures. The findings revealed significant enhancements in positional accuracy through camera calibration procedures, especially with pre-existing calibration models. 

\subsection{Evaluation of the quality}\label{subsec2}

The assessment of the quality of the 3D modeling and the definition of the possible usability of the generated products are necessary for the determination of the efficiency of the configuration used. As discussed in \cite{garcia2021}, the RMSE value, calculated by Equation 1, has been employed for error analysis and prediction accuracy evaluation: 

\begin{equation}
RMSE=\sqrt{\frac{\sum_{i=1}^{n}\left(E{i}-R{i}\right)^{2}}{n}}\label{eq1}
\end{equation}where n is the number of samples, E\textsubscript{i} is the estimated value at position i, and R\textsubscript{i} is the value measured at position i.
These procedures assess deviations between measured and actual elements to delineate variations more accurately. When evaluating 3D model quality, estimated values stem from computational modeling data, while actual values are derived from precise field measurements \cite{segantine2015}.

Analyzing these values, along with Quality of Adjustment values from processing software, enables a thorough evaluation of the entire 3D reconstruction process, covering specific product regions. This helps pinpoint and address errors in capture and modeling techniques, enhancing data precision and reliability.

\section{Materials and Method}\label{sec3}
It is important to understand that the primary objective of this study is to evaluate the possible use of the SfM technique for achieving submillimeter measurements. Therefore, at this stage of the experiments, we are not considering external influences such as the object's texture and the lighting conditions in the capture area.

We utilized the commercial application Agisoft Metashape Professional (version 1.8.5), a widely used blackbox computational solution in SfM works, for the 3D modeling software. Table~\ref {tab1} presents the configurations for the workflow we employed for processing. This study focuses on evaluating the technique's effectiveness by improving quality settings, without discussing the workflow settings. For details, refer to \cite{leon2015, james20173, tinkham2021}.

\begin{table}[h!]
    \caption{Workflow configurations used in all model processing}\label{tab1}
    \centering
    \begin{tabular}{|c|c|c|c|}
        \hline
        \multicolumn{4}{|c|}{\textbf{Workflow Configuration}} \\    \hline
        \multicolumn{2}{|c|}{\textbf{Allign Photos}} & \multicolumn{2}{c|}{\textbf{Mesh}} \\     \hline
        Accuracy & \textit{High} & Quality & \textit{High} \\     \hline
        Key Point limit & \textit{60,000} & Face Count. & \textit{High} \\     \hline
        Tie Point & \textit{0} & Depth Filtering & \textit{Aggres.} \\     \hline 
        \multicolumn{2}{|c|}{\textbf{Dense Cloud}} & \multicolumn{2}{c|}{\textbf{Build Tiled Model}} \\     \hline
        Quality & \textit{High} & Quality & \textit{High} \\    \hline
        Depth Filtering & \textit{Aggres.} & Face Count. & \textit{High} \\     \hline
       \multicolumn{2}{|c|}{\textbf{}} & Depth Filtering & \textit{Aggres.} \\     \hline
    \end{tabular}
\end{table}

\subsection{Region of Interest}\label{subsec3}
The experiments were conducted at the Department of Transportation Engineering of the São Carlos School of Engineering at the University of São Paulo. The room used has natural lighting and artificial lights that influenced the camera settings throughout the capture process.

In this study, we opted to simplify photographic capture stages by replicating a specimen used in structural testing. We employed a wooden board measuring 210 cm x 80 cm x 4 cm, deliberately fitted with metal rulers of varying lengths at strategic positions in two separate areas. These targets functioned as both Scale Bars (SBs) and Check Bars (CB), facilitating the SfM modeling process.

During analysis, the area of the board, designated as the region of interest, was divided into three proportional sectors. A portion measuring 175 cm x 46 cm represented the laboratory test specimen, while sets of Scale Bars (SBs) were randomly placed in each sector outside this area. These eight sets, varying in length and orientation, were positioned relative to the board's axes.

\begin{figure}[h!]
\centering
\includegraphics[width=\textwidth]{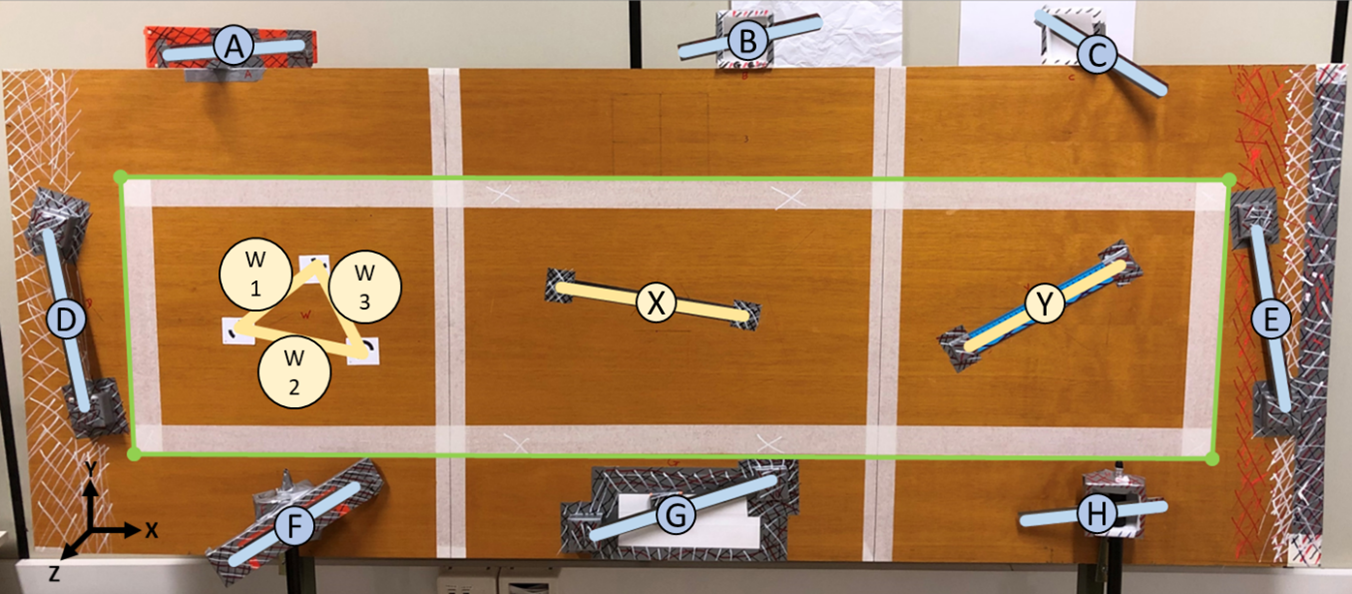}
\caption{Distribution of the SBs along the edges of the sectors within the region of interest, while CBs were positioned across the area of the analyzed object}\label{fig1}
\end{figure}

CBs were placed within each sector of the wooden board, specifically in the internal areas representing the specimen. Figure~\ref{fig1} depicts the distribution of SBs in connection with the sectors of interest and the specimen, along with the positioning of CBs relative to the analyzed object.
\subsection{SB Distribution}\label{subsec4}
As highlighted in \cite{westoby2012}, for endeavors focused on monitoring large-scale structures, optimal gains in positional accuracy of 3D modeling necessitate a strategic distribution of targets tailored to the requirements and constraints of the task at hand. This includes employing sets of targets with varied height settings and patterns.

Therefore, three models were developed for the distribution of SBs, encompassing six distinct configurations, each initially featuring three bars. The number of bars was subsequently incrementally raised until reaching a total of eight SBs. Table~\ref{tab2} shows the distribution configurations of SBs used because of the variation in the number of reference scale bars used.

\begin{table}[h]
\caption{Configuration of the distribution sets for the quality analysis because of the number and position of the Scale Bars}\label{tab2}
    \centering
    \begin{tabular}{|c|c|c|c|c|c|}
    \hline
    \textbf{SB} & \textbf{D01} & \textbf{SB} & \textbf{D02} & \textbf{SB} & \textbf{D03} \\
    \hline
    \textbf{3} & A-E-G & \textbf{3} & B-F-G & \textbf{3} & B-D-E \\
    \hline
    \textbf{4} & A-D-E-G & \textbf{4} & B-C-F-H & \textbf{4} & B-D-E-G \\
    \hline
    \textbf{5} & A-D-E-G-H & \textbf{5} & B-C-F-G-H & \textbf{5} & B-D-E-G-H \\
    \hline
    \textbf{6} & A-B-D-E-G-H & \textbf{6} & A-B-C-F-G-H & \textbf{6} & A-B-D-E-G-H \\
    \hline
    \textbf{7} & A-B-D-E-F-G-H & \textbf{7} & A-B-C-E-F-G-H & \textbf{7} & A-B-D-E-F-G-H \\
    \hline
    \textbf{8} & A-B-C-D-E-F-G-H & \textbf{8} & A-B-C-D-E-F-G-H & \textbf{8} & A-B-C-D-E-F-G-H \\
    \hline
    \end{tabular}
\end{table}
The initial distribution, designated as D01, prioritized the allocation of SBs in specific sectors of the object of interest. Subsequently, additional bars were introduced to attain a more balanced distribution of reference elements. The goal of D01 was to analyze the distribution impact of SBs, with an emphasis on achieving a symmetrical distribution across the analyzed regions.

In the second distribution (D02), initial SBs were used, deliberately omitting elements on the vertical edges of the area of interest. Later, additional horizontal reference elements were introduced while preserving this distribution. Finally, SBs on the vertical edges were added, allowing assessment of a modeling scenario where using reference elements in certain object regions is impractical.

The third distribution (D03) adopted a configuration in which the central reference elements were prioritized and, subsequently, the SBs present on the edges of the analyzed regions of interest. D03 sought to assess the influence of prioritizing central SBs as opposed to the reference elements at the vertices of the region of interest, to comprehend the impact of this approach on the analysis.

\subsection{Data Acquisition}\label{subsec5}
\begin{figure}[h]
\centering
\includegraphics[width=\textwidth]{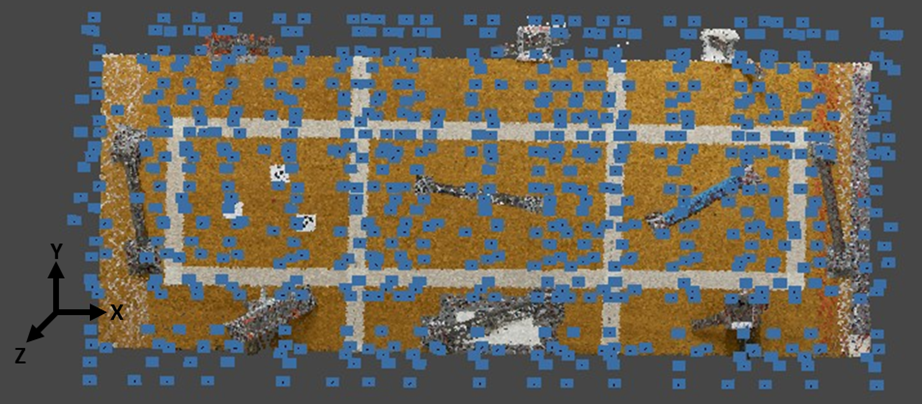}
\caption{Processing conducted representing the point cloud (in the background) and the positioning of the cameras at the time of capture (blue squares), in this processing, all images (vertical and oblique) are displayed.}\label{fig2}
\end{figure}

We selected specific camera settings to obtain images with minimal noise (ISO 100) and full object focus (f/11), with a focal length of 35 mm. Accordingly, to maintain the proper balance of the Exposure Triangle, we adjusted the shutter speed to ensure high illumination (D+1). The photographic captures were conducted by systematically varying the camera positioning for each of the work analyses, establishing a capture method akin to a regular vertical grid model, as presented in Figure~\ref{fig2}.

\begin{figure}[h]
    \centering
	\begin{subfigure}[b]{0.47\textwidth}		
		\centering
		\includegraphics[width=\textwidth]{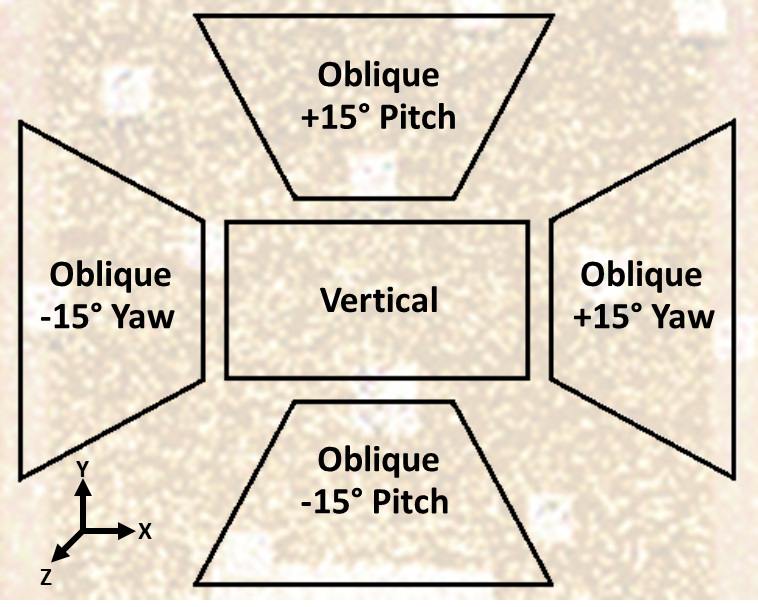}
		\caption{}
		\label{fig:sub:subfiga}
	\end{subfigure}
	\begin{subfigure}[b]{0.47\textwidth}		
		\centering
		\includegraphics[width=\textwidth]{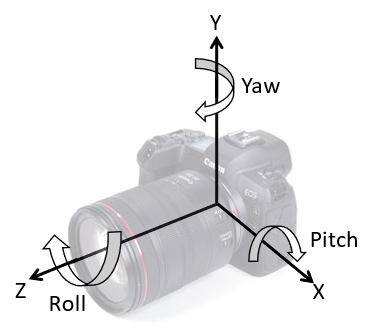}
		\caption{}
		\label{fig:sub:subfigb}
	\end{subfigure}
	\caption{a) The represented capture set used vertical and oblique images, with rotation of ± 15° in Yaw, followed by captures with ± 15° in Pitch; (b) Representation of possible rotation movements of the camera.}
	\label{fig:detectionresults}
\end{figure}

In Figure~\ref{fig:sub:subfiga}, a depiction of the photographic capture is presented, consisting of a vertical image of the wooden board and sets of oblique images. The oblique images were initially acquired with a camera rotation of ± 15° in the Yaw direction (Y-axis) and subsequently with a rotation of ± 15° in the Pitch direction (X-axis). No rotation movements in the camera's Roll direction (Z-axis) were executed for either set of captures. Figure~\ref{fig:sub:subfigb} illustrates the axes associated with potential camera rotation movements.

To evaluate the feasibility of using the SfM technique in structural tests, we selected a capture distance of 1 meter to ensure safety during laboratory experiments. This distance also maintains an efficient Ground Sample Distance (GSD) of 0.15 mm, which represents the distance between two consecutive pixels on the ground or surface of the object. Images were captured using a Canon EOS R, featuring a resolution of 6,720 x 4,480 pixels, on a full-frame sensor (35 mm x 24 mm).

As demonstrated in \cite{nesbit2019} and \cite{james20173}, in 3D modeling research involving the capture of objects at distances of several meters, a combination of images taken vertically and obliquely to the specimen was utilized to enhance the positional quality of three-dimensional modeling. Furthermore, this combination provides a model representation with a higher level of detail of the objects of interest, especially for objects with complex geometry.

To determine the lengths of the sets of SBs and CBs, which were employed for scaling and checking the developed 3D models, a Starrett Digital Caliper from the EC799A-8 series was utilized. This equipment demonstrates a degree of accuracy of ± 0.02 mm for measurements up to 10 cm and an accuracy of ± 0.03 mm for measurements exceeding 10 cm \cite{Starret}. 

\begin{figure*}[t]
\centering
\includegraphics[width=\textwidth]{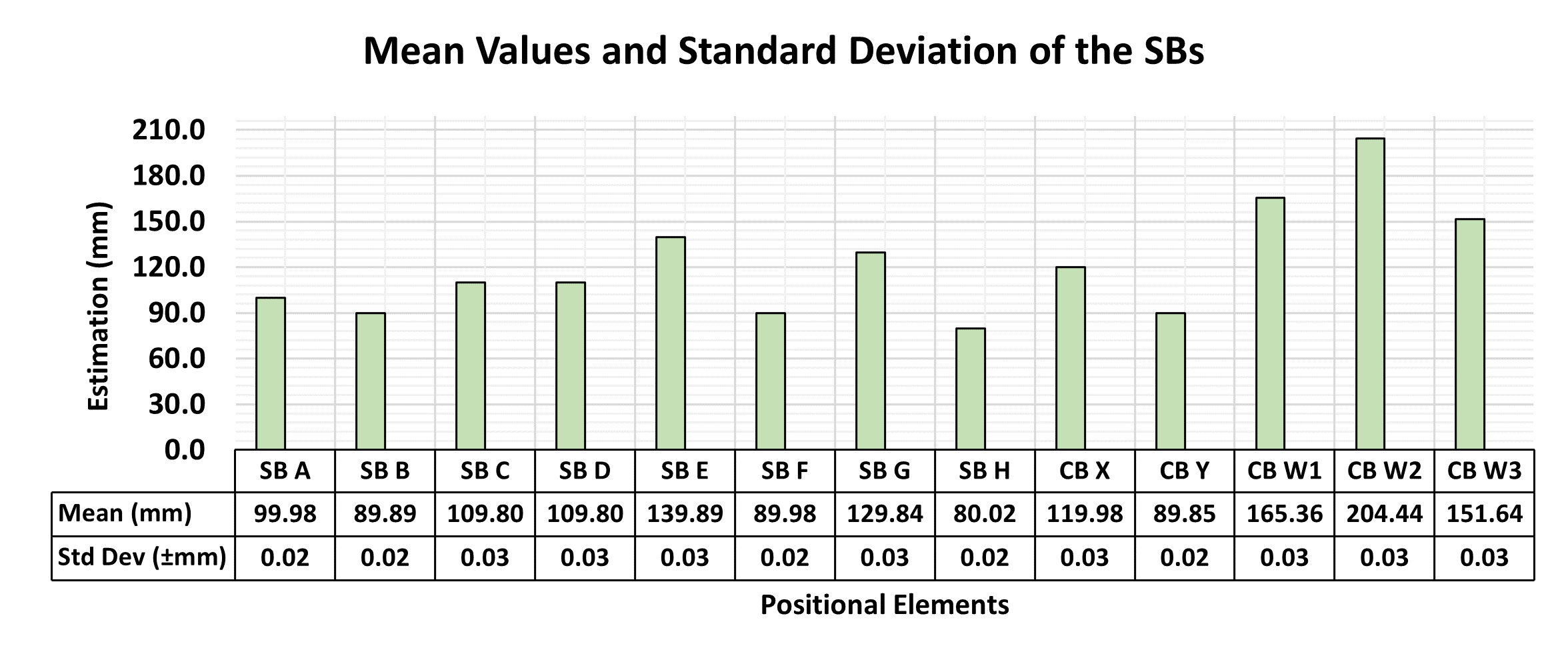}
\caption{Representation of the mean lengths and Standard Deviation of the measurements of each positional element, obtained through a series of 5 measurements.}\label{fig4}
\end{figure*}

Each positional element underwent five measurements by the same operator. The accuracy of reference elements and the SfM modeling's positional precision will significantly influence the potential precision of the models developed in this research. Figure~\ref{fig4} depicts accuracy estimates, measured in hundredths of a millimeter, for each positional bar used in the study.

\subsection{Camera Calibration Models}\label{subsec6}
Because of the current popularization of SfM modeling, several camera calibration modes with user-friendly procedures can be performed effectively. In this study, because of the characteristics of the analyses carried out, we chose to use three different models, On-the-Field Calibration (OtF) and the pre-calibration models known as Computer Vision Calibration (CVC) and Pre-Self Calibration (PSC).

Unlike the other models used, the OtF calibration model, the parameters are estimated throughout the 3D coordinate determination process, during the SfM stage, automatically, through the Bundle Adjustment process.

For the CVC calibration model, as discussed in \cite{luhmann2023}, a planar chessboard object was photographed from various angles to estimate camera behavior. In this study, we took fifty images of this calibration object and used the library OpenCV 4.6.0 to determine the camera parameters.

Due to the PSC calibration model making use of a test field with several SBs, which must have similar characteristics to the main capture environment, we decided to carry out a modeling process of the region of interest itself, analyzed in this work. Nine SBs were employed, with eight used experimentally and the central CB selected. The varied bar configurations aimed for a comprehensive representation of camera characteristics during capture, as discussed in \cite{garcia2021}.

\subsection{Quality Assessment}\label{subsec7}
SfM generates a camera-referenced point cloud, requiring scaling for positional accuracy\cite{hartley2003}. Various SB formats and configurations with known lengths were employed to size and refine the products. In the initial phase, 3D point positional accuracy was assessed by comparing virtual model-derived bar lengths with the reference measurements using a digital caliper.

The Metashape software provides numerical values extracted from the diagonal of the covariance matrix to facilitate a thorough analysis of 3D modeling adjustments, visually depicted through a color gradient. This representation derived from the covariance matrix diagonal assists in gauging the precision of parameter estimates and pinpointing areas requiring refinement or indicating uncertainty within SfM models \cite{agisoft185}.

The primary objective of this study was to ascertain the intrinsic quality of 3D modeling using Agisoft Metashape software and to assess its viability for laboratory test evaluations. The evaluation of modeling quality for each experiment involved analyzing RMSE values, correlating the lengths of real and virtual CBs, and examining quality indicators for software-rendered 3D model adjustments.

\section{Results}\label{sec4}
\subsection{Camera Calibration Models}\label{subsec8}
\begin{figure*}[t]
    \centering
    \includegraphics[width=\textwidth]{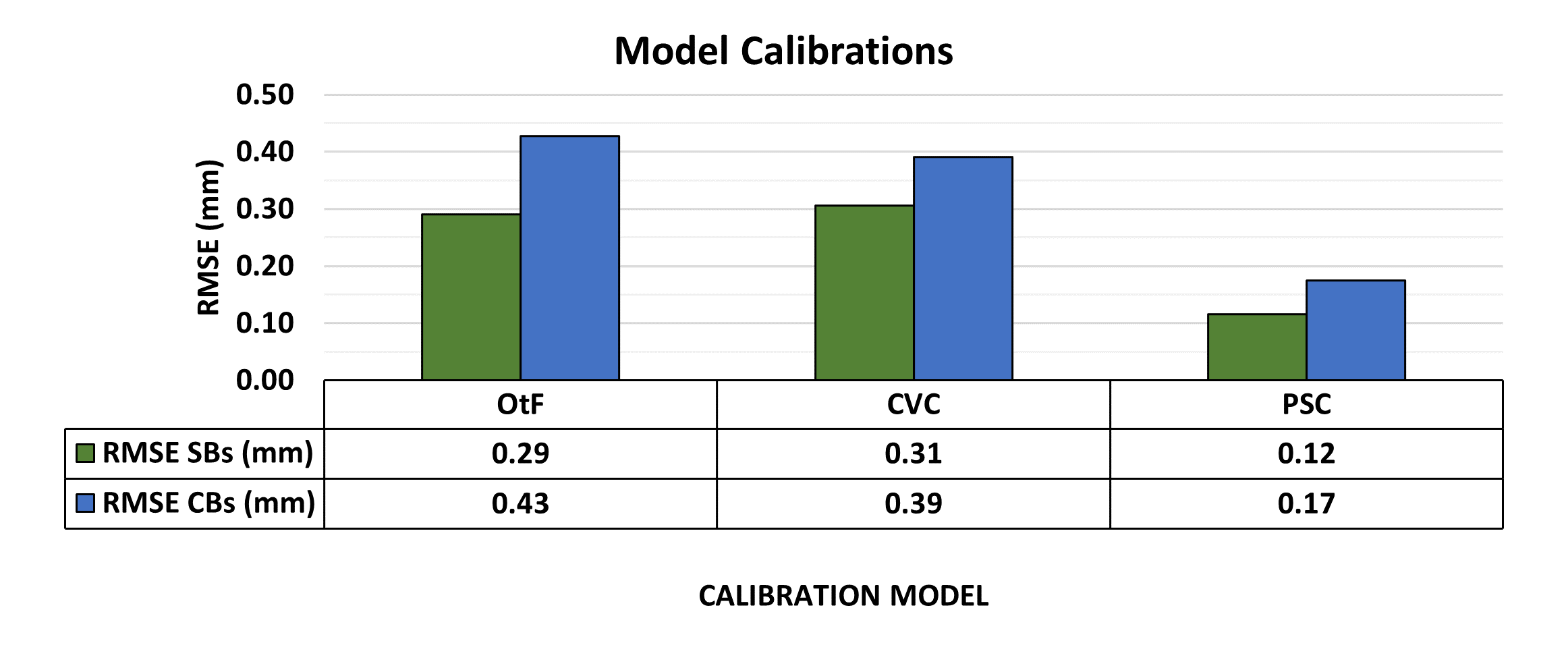}
    \caption{RMSE values obtained for 3D modeling with three different camera calibration models, using sets of images of the test piece with 80\% overlap and employing the maximum number of SBs and CBs, showing a significant advantage for the product that utilized the pre-calibrated PSC model.}\label{fig5}
\end{figure*}

The activities to analyze the quality levels of the regions of interest, for each of the calibration models, were developed using images with an overlap of 80\%, with the maximum number of SBs (8) and, for the pre-calibrated models, the respective calibration parameters. The values obtained from the RMSE for each modeling performed are presented in Figure~\ref{fig5}.

The 3D modeling procedures employed with the OtF and CVC calibration models exhibited comparable quality levels. Nevertheless, a slight advantage was observed for the computer vision method when compared to the automatic OtF model. This superiority can be attributed to the fact that the CVC technique provides pre-calibrated camera parameters, obtained through a series of photographic captures of a flat checkered board, thereby streamlining the modeling process.

Nonetheless, it is crucial to emphasize that this advantage is constrained to a slight difference, due to the nature of the structure of the object employed in the pre-calibration process. Specifically, when utilizing a planar calibration object, the CVC model adversely affects depth information in the modeling process, as demonstrated in \cite{garcia2021}. On the other hand, OtF takes advantage of the three-dimensional scene of interest to derive relevant camera parameters, which results in modeling with similar quality to the CVC model.

In turn, the PSC model, using a three-dimensional environment resembling the scene of interest in its pre-calibration process, demonstrated the best quality, with a value of 0.17 mm in RMSE of CBs. Similar to the preceding models, the results obtained with the PSC model achieve submillimeter standards for both quality values. Notably, there is a significant improvement of approximately 40\% compared to the other analyzed models. 

The values obtained by PSC were adopted as a reference for an evaluation of the calibration parameters between pre-calibrated models and those from modeling because of the superior performance of the model in terms of quality values in the analysis. 

The calibration parameters under consideration encompassed the following components: the principal distance (c) which represents the focal length; the coordinates of the principal point (xp and yp) signifying the offset distance from the center of the sensor; radial distortion parameters (K1, K2, and K3) arising from lens imperfections; and decentering distortion coefficients (P1 and P2) originating from deviations in optical component alignment or centering. Table~\ref{tab3} shows the calibration parameters, revealing a minor disparity between OtF and the reference values for all parameters.

\begin{table}[h]
    \caption{Camera calibration parameters are estimated for each of the models analyzed in this research}\label{tab3}
    \centering
    \begin{tabular}{|>{\centering\arraybackslash}p{0.6cm}|>{\centering\arraybackslash}p{0.7cm}|>{\centering\arraybackslash}p{0.7cm}|>{\centering\arraybackslash}p{0.8cm}|>{\centering\arraybackslash}p{1.14cm}|>{\centering\arraybackslash}p{1.24cm}|>{\centering\arraybackslash}p{1.14cm}|>{\centering\arraybackslash}p{1.24cm}|>{\centering\arraybackslash}p{1.14cm}|}
        \hline
        \textbf{} & \textbf{c (mm)}    & \textbf{xp (mm)}  & \textbf{yp (mm)}   & \textbf{K1}     & \textbf{K2}        & \textbf{K3}       & \textbf{P1}        & \textbf{P2}       \\ \hline
        \textbf{OtF}   & 35.487    & 0.360    & -0.063    & 7.35E-05    & -9.78E-07 & 6.39E-11 & -3.32E-05 & 3.20E-05 \\ \hline
        \textbf{CVC}   & 35.511    & 0.361    & -0.064    & 7.37E-05    & -9.78E-07 & 8.47E-11 & -3.10E-05 & 3.12E-05 \\ \hline
        \textbf{PSC}   & 35.500    & 0.359    & -0.064    & 7.32E-05    & -9.76E-07 & 9.12E-11 & -3.08E-05 & 3.15E-05 \\ \hline
    \end{tabular}
\end{table}

Although the calibration parameters exhibited similarities, notable differences were observed for the lens distortion parameters (K3 and both Px) in the CVC model. This discrepancy may be associated with the calibration method involving a planar object and variations in the photographic capture process of the object.

Finally, despite the relative proximity of some calibration parameters, the RMSE values obtained demonstrated the advantage of using the PSC model, to others, in 3D coordinate estimation processes with a capture distance of up to 1 meter.

\subsection{Comparison between different SB clusters}\label{subsec9}
Based on previous results, that underscored the substantial benefits associated with the utilization of the PSC calibration model, we proceeded to initiate 18 distinct 3D modeling procedures. Each of these procedures integrated the calibration model in question. Moreover, we varied the distribution configurations and quantities of SBs for each of these processes, according to the evaluations necessary for this paper.

In total, three sets of distributions were elaborated, denoted as Distribution 01, Distribution 02, and Distribution 03. These sets were designed to prioritize distinct layouts of SBs, to simulate diverse requirements and methodologies for potential structural tests. Each set consisted of six configurations about the quantity and arrangement of SBs.

\begin{figure}[h!]
    \centering
    \includegraphics[width=\textwidth]{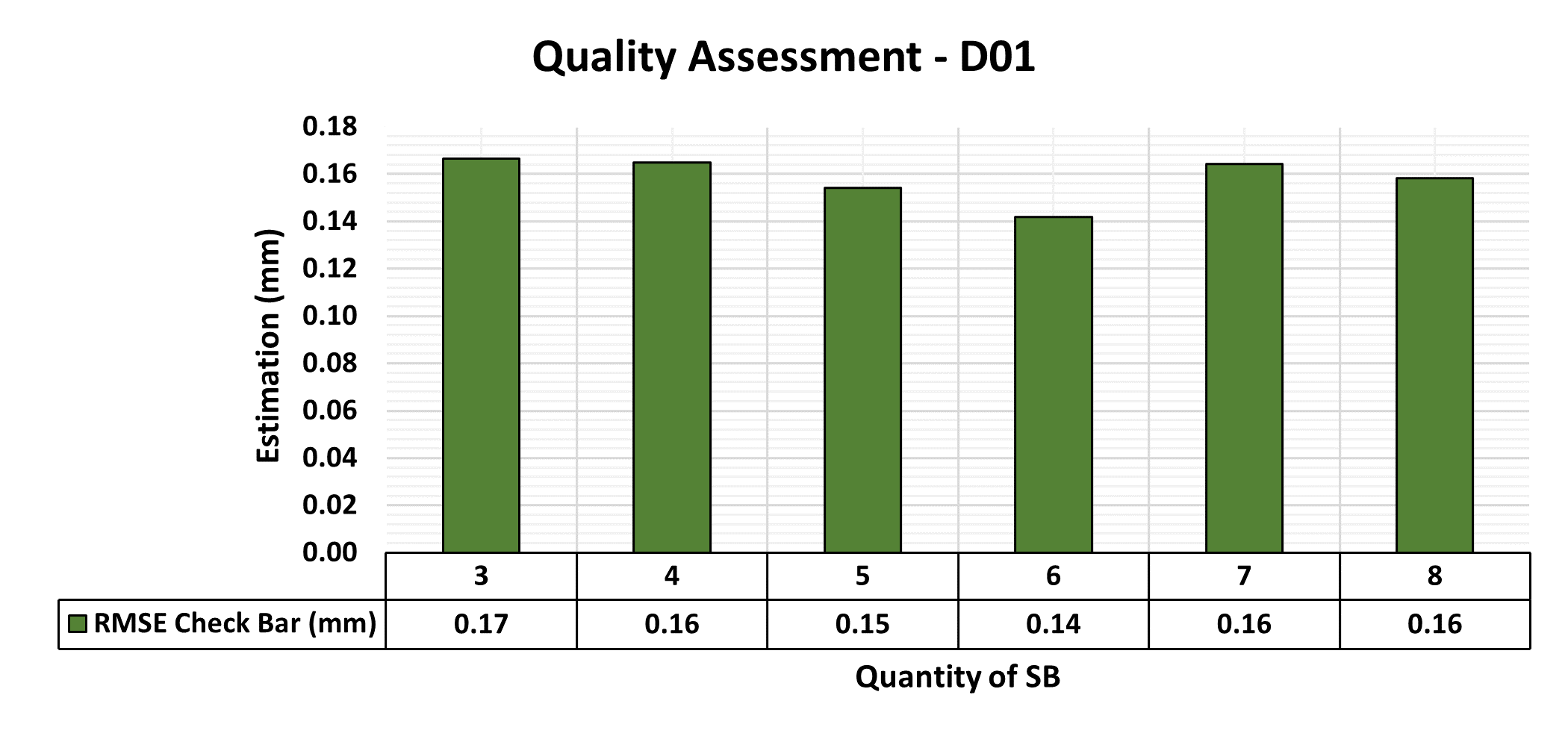}
    \caption{RMSE values of CBs for different SB dispersion configurations in Distribution 01 (D01) set. For D01, consistency around the value of 0.16 mm is evident.}\label{fig6}
\end{figure}

Figure~\ref{fig6} displays the RMSE values of Distribution 01, focusing on the incorporation of SBs to achieve a more uniform distribution along the edges of the object. This distribution yields notably balanced RMSE values for all combinations, averaging around 0.16 mm. 

\begin{figure}[h!]
    \centering
    \includegraphics[width=\textwidth]{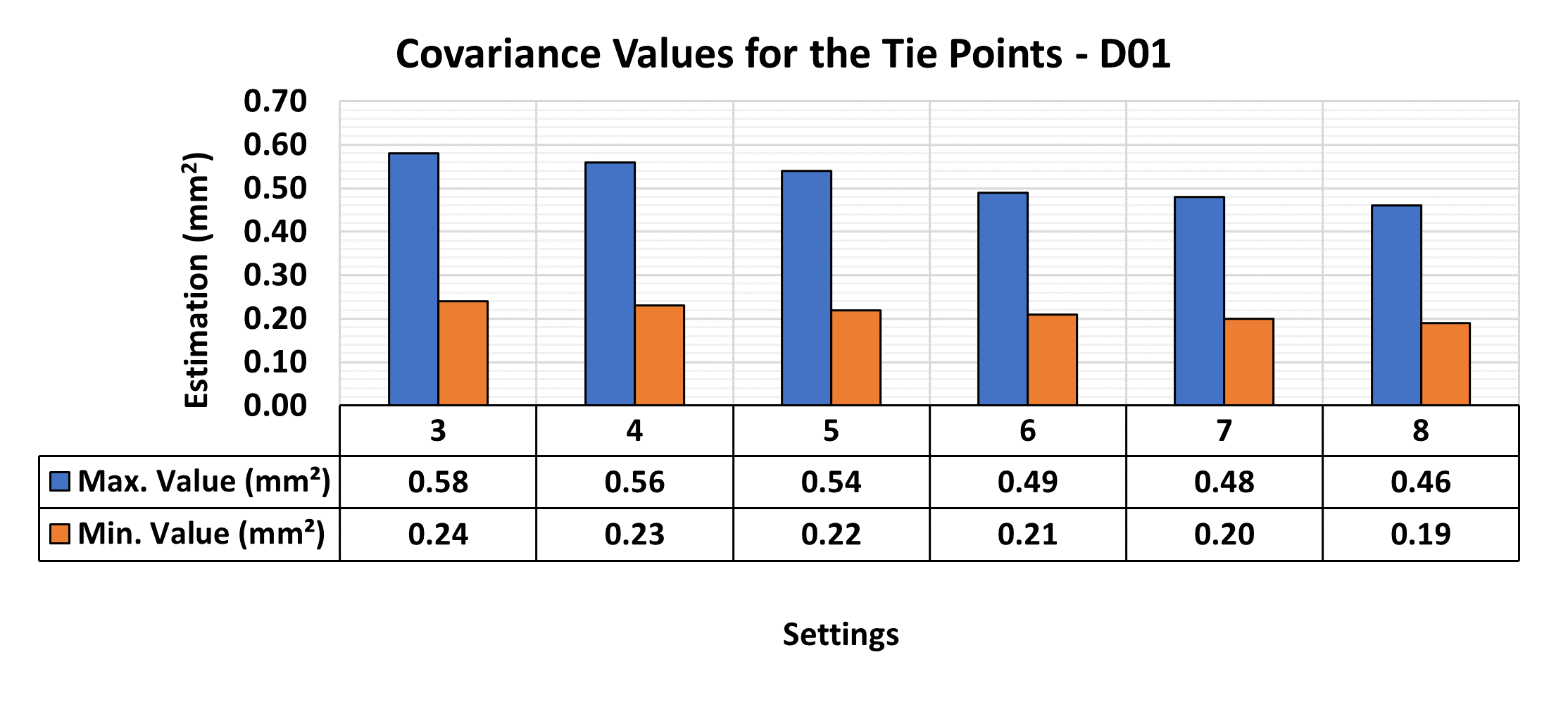}
    \caption{RMSE values of CBs for different SB dispersion configurations in Distribution 01 (D01) set. For D01, consistency around the value of 0.16 mm is evident.}\label{fig7}
\end{figure}

However, as depicted in Figure~\ref{fig7}, an improvement in terms of Covariance Matrix values for each modeling set is evident with the inclusion of more reference elements in the modeling process, with the optimal result achieved using 8 SBs.

\begin{figure}[h]
    \centering
    \includegraphics[width=\textwidth]{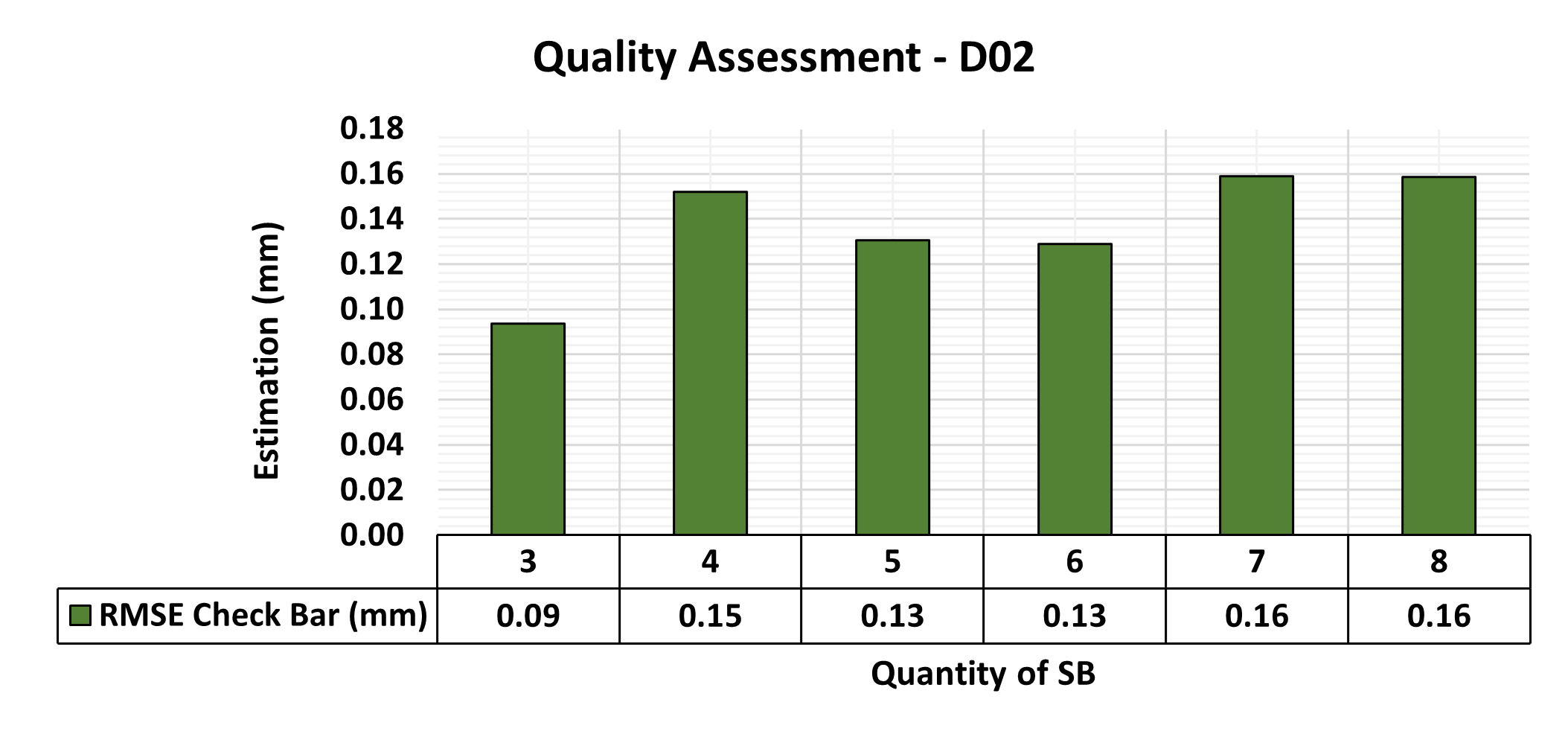}
    \caption{RMSE values of CBs for different SB dispersion configurations in Distribution 02 (D02) set. In D02, a value of 0.09 mm is observed for 3 SBs, while consistency around 0.14 mm is maintained for the other configurations.}\label{fig8}
\end{figure}

Distribution 02, as presented in Figure~\ref{fig8}, initially prioritized incorporating reference elements along the horizontal edges, thereby avoiding the use of SBs on the vertical edges of the region of interest. The best RMSE was obtained with the initial set of four SBs, while the subsequent results consistently demonstrated behavior following the varying quantity of SBs, maintaining an average value of 0.14 mm. 

\begin{figure}[h!]
    \centering
    \includegraphics[width=\textwidth]{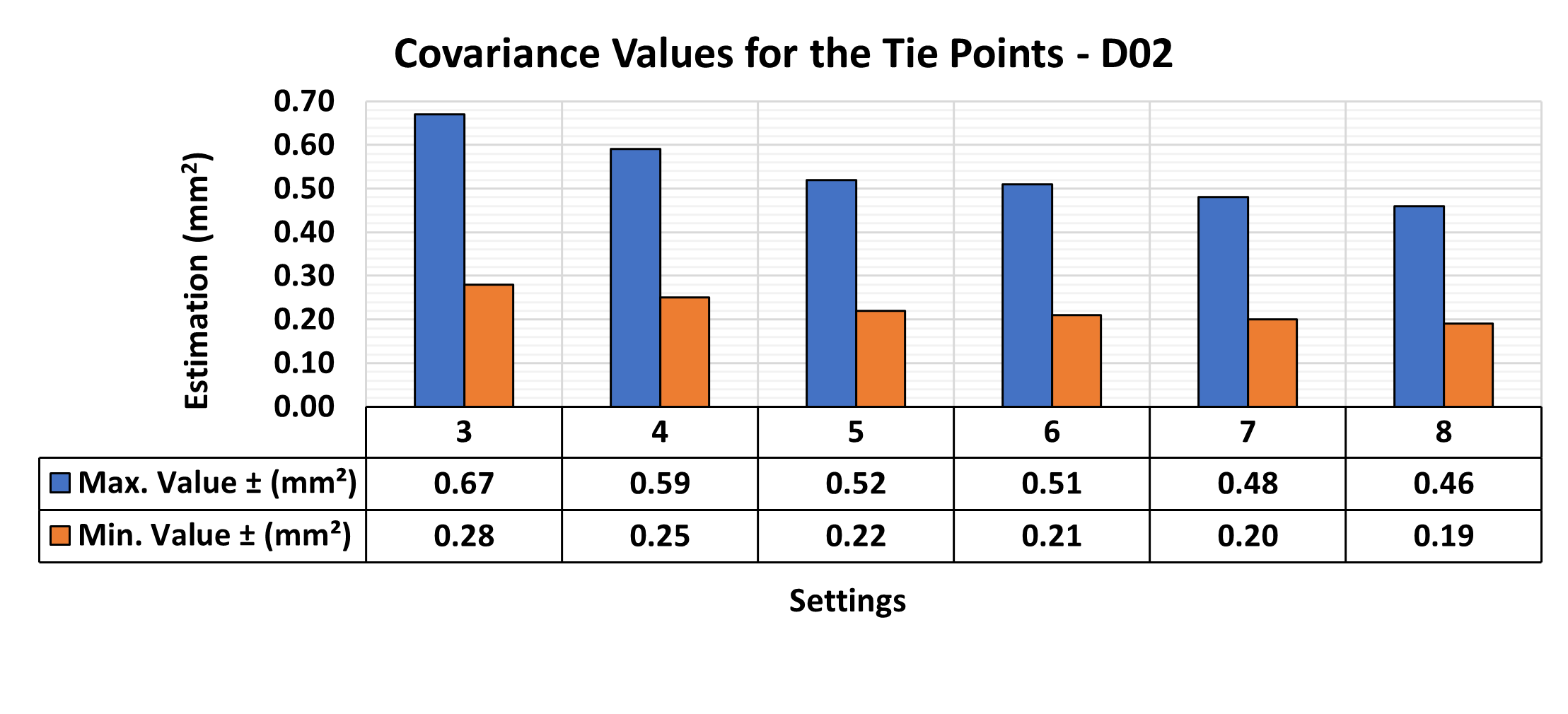}
    \caption{Maximum and minimum values obtained from the Covariance Matrix indicate the quality of adjustment for each 3D modeling set. In the case of D02, superior results were noted for the configuration with 8 SBs, yielding a maximum value of 0.46 mm² and a minimum of 0.19 mm², as in D01.}\label{fig9}
\end{figure}

Conversely, as presented in Figure~\ref{fig9}, the values presented by the software regarding the Covariance Matrix for each modeling set exhibited a behavior like that observed in D01, where the least favorable modeling was obtained with the initial set of 3 SBs, and a quality improvement was observed with the inclusion of additional reference elements.

\begin{figure}[h!]
    \centering
    \includegraphics[width=\textwidth]{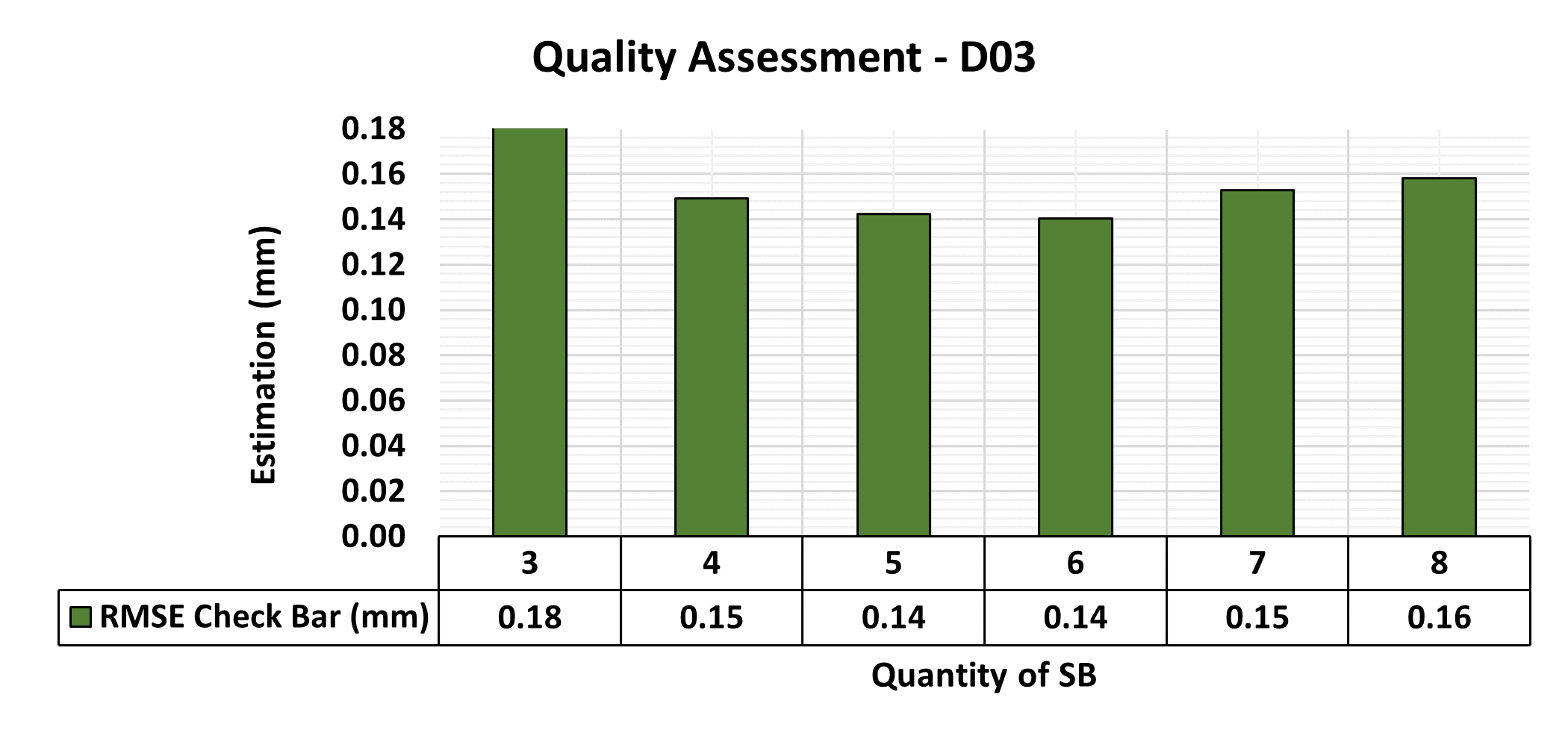}
    \caption{RMSE values of CBs for different SB dispersion configurations in Distribution 03 (D03) set. For D03, consistency around the value of 0.15 mm is evident.}\label{fig10}
\end{figure}

The third distribution, presented in Figure~\ref{fig10}, emphasized utilizing SBs positioned in the central sections of the edges in the region of interest. The remaining elements were randomly added to the entirety of SBs. Similar to D01, this set of distributions exhibited consistent RMSE values, regardless of the number of SBs used, with an average value of approximately 0.15 mm, mirroring the other sets employed in this study.

\begin{figure}[h!]
    \centering
    \includegraphics[width=\textwidth]{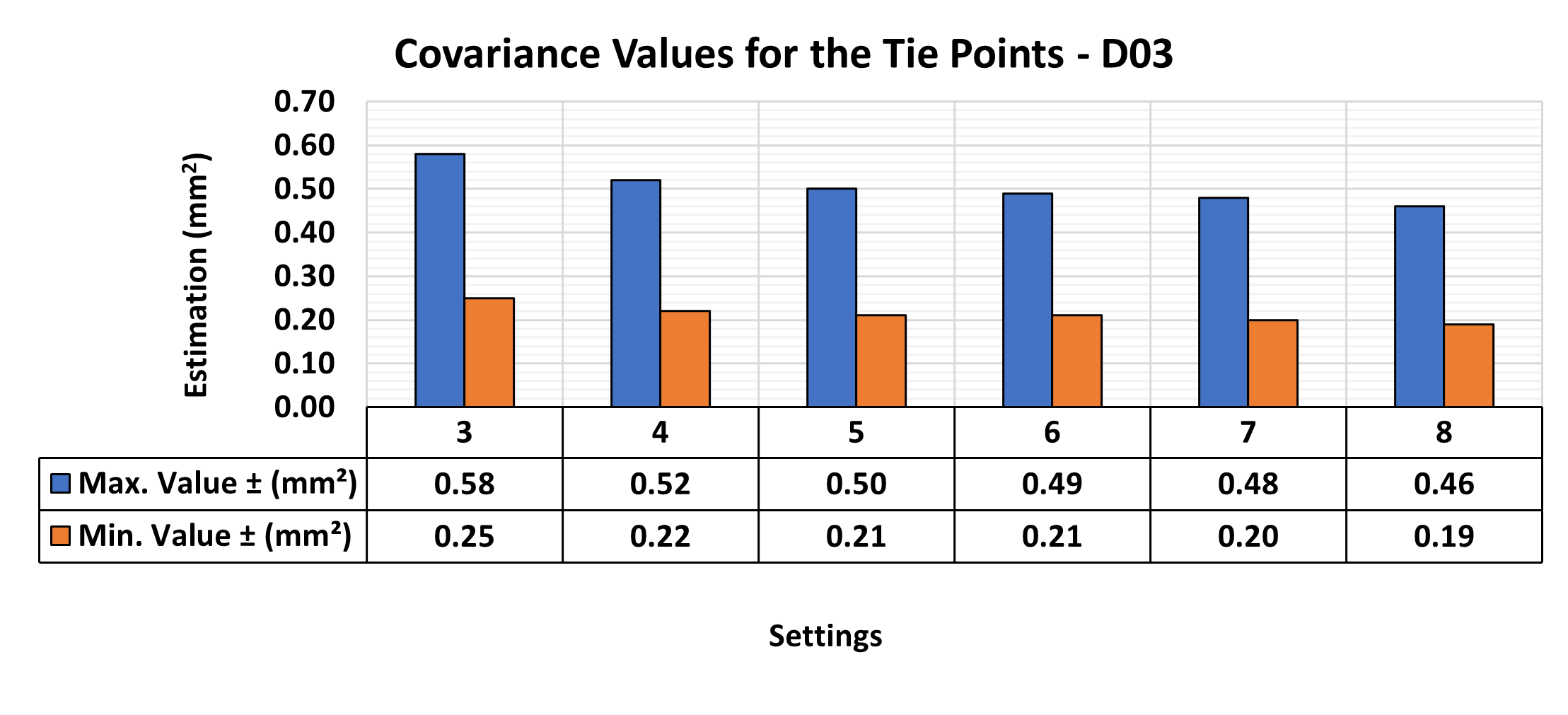}
    \caption{Maximum and minimum values obtained from the Covariance Matrix indicate the quality of adjustment for configuration D03. As observed in the other analyses, the increase of SBs leads to improved values, as demonstrated in the maximum and minimum values of the Covariance Matrix.}\label{fig11}
\end{figure}

Figure~\ref{fig11} presents, as observed in the analysis of other distributions, the Covariance Matrix values improved as more SBs were incorporated, ranging from 0.58 mm² to 0.46 mm² for the maximum value and from 0.28 mm² to 0.19 mm² for the minimum value. The least favorable result was obtained with the initial set, while the most favorable was achieved with the inclusion of all available SBs.

The three sets of distributions exhibited consistency in both values and behavior of the RMSE as additional elements were incorporated into the modeling. Notably, configuration D02, with the initial set of SBs, stood out by yielding the lowest RMSE value in the entire analysis (0.09 mm). 
However, regarding the values obtained from the Covariance Matrix for the modeling, this configuration yielded a result significantly higher than the others in terms of maximum values (0.67 mm²), leading to its exclusion from further analyses.

Despite the positional quality remaining consistent across various SB quantity settings, there was a significant improvement in the values related to the Covariance Matrix of the modeling with the gradual addition of more bars. This assertion finds support through a comparative analysis of the results obtained concerning the values from the Covariance Matrix of all modeling sets.

This observed improvement can be attributed to the enhanced precision available for representing the diverse regions of the analyzed object. The utilization of a greater number of SBs led to a more precise positional representation of the region of interest, empowering the 3D modeling process to gather additional information for refining the produced products with increased efficiency.

Therefore, based on this analysis, it can be concluded that, for each verified distribution, the best outcome was achieved when utilizing the maximum number of SBs. This configuration ensured a more comprehensive coverage of the entire contour of the region of interest. It must be adopted for short-distance 3D modeling activities with a focus on structural tests.

However, in certain laboratory experiments, the placement of many SBs throughout the entire surroundings of the object may be impractical due to the arrangement of other devices, potentially resulting in areas with occluded information in the structural analysis. It is the user's responsibility to identify the specific demands of the tests and analyses to be conducted and distribute the SBs properly throughout the entire surroundings of the object, according to the specific needs of the experiment.

\subsection{Analysis of Variation in the Overlap Percentages}\label{subsec10}
Studies such as the one by \cite{gerke2016}, although focused on photographic capture using UAV, performed similar evaluations with different overlapping values, obtaining the best results using higher overlapping rates. However, given the variations in collection methods and modeling objectives, specific analyses were undertaken for the modeling environment at extremely short distances. From the more efficient results obtained, using the PSC calibration and distribution with 8 SB, new sets of processing were carried out aiming to determine the influence of different overlapping on the quality of the 3D coordinate estimation.

In this study, a range of overlapping percentages from 60\% to 80\% was employed, and while the use of 90\% is possible, it would result in a substantial set of images, and the positional gains may not be acceptable to the time invested in this configuration. Table~\ref{tab4} displays various configurations of overlap settings and the corresponding image quantities for each 3D model.

\begin{table}[h]
    \caption{Configurations for each combination of overlay (horizontal and vertical) and the number of images utilized for generating each 3D model}\label{tab4}
    
    \begin{tabular}{|>{\centering\arraybackslash}p{3cm}|>{\centering\arraybackslash}p{3cm}|>{\centering\arraybackslash}p{3cm}|}
    \hline
    \% \textbf{Horizontal} & \% \textbf{Vertical} & \textbf{Number of Images} \\ \hline
    60            & 60          & 68               \\ \hline
    70            & 70          & 102              \\ \hline
    80            & 80          & 153              \\ \hline
    \end{tabular}
\end{table}

\begin{figure}[h!]
    \centering
    \includegraphics[width=\textwidth]{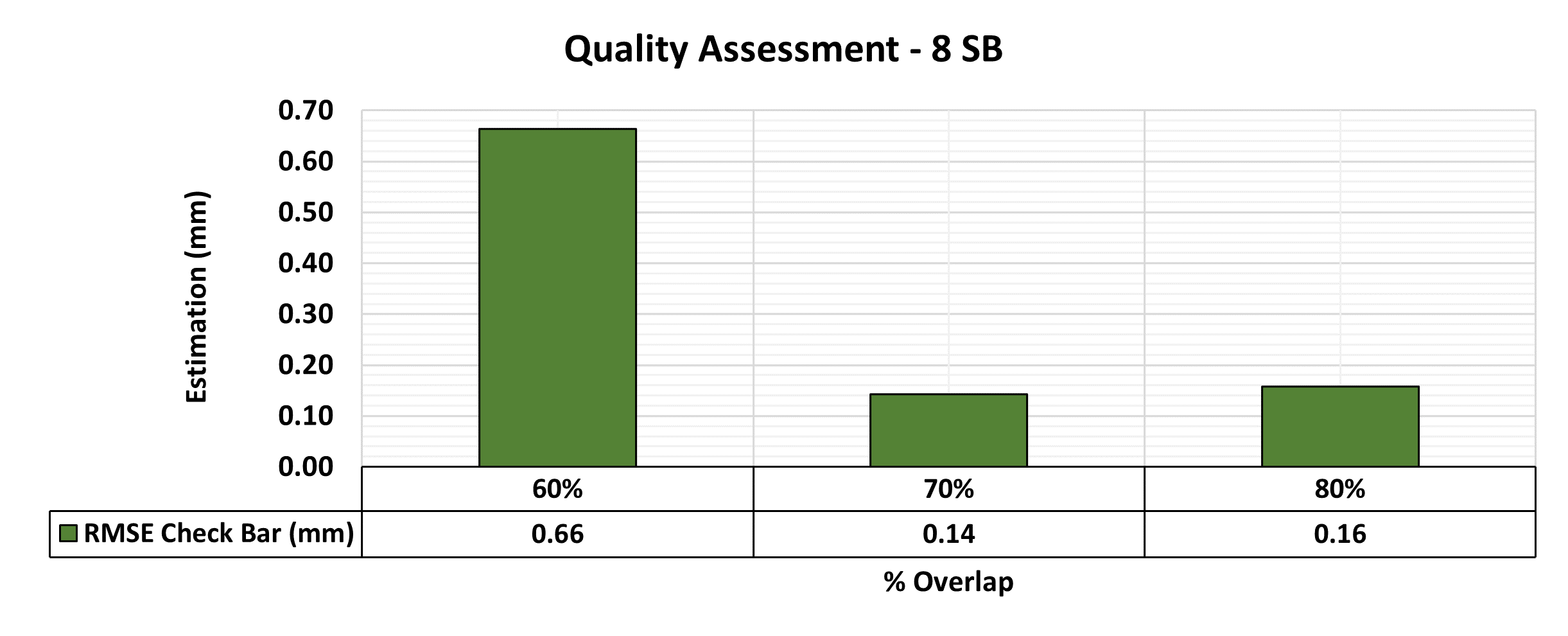}
    \caption{RMSE values obtained for different overlap settings. The best result was achieved with the maximum number of SBs (8) and a 70\% overlap configuration (0.14 mm), followed by the 80\% overlap set (0.16 mm), while the worst result was obtained for the 60\% overlap configuration (0.66 mm).}\label{fig12}
\end{figure}

Based on previous experiments conducted in our laboratories, where no improvements were observed in terms of positional quality, the use of a distinct overlap ratio for each axis (horizontal and vertical) was discarded. Therefore, the configurations of overlap were kept constant for both axes, varying from 60\% to 80\%. Figure~\ref{fig12} presents the outcomes concerning the quality assessment for each set of images with different overlapping rates.

\begin{figure}[h!]
    \centering
    \includegraphics[width=\textwidth]{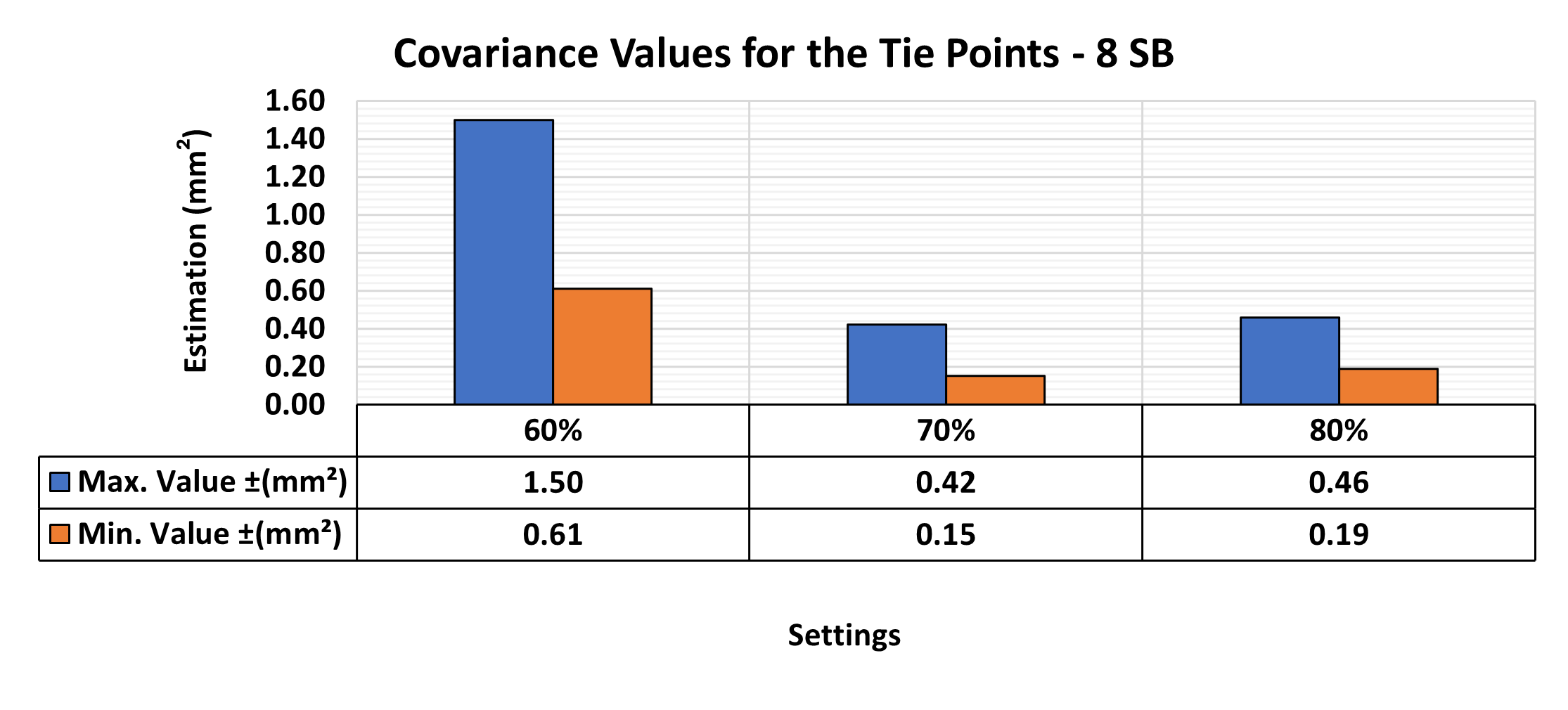}
    \caption{Maximum and minimum values obtained from the Covariance Matrix from the different overlap ratios. Similar to the RMSE values, a consistent pattern was observed in the Covariance Matrix values, with the 60\% overlap ratio resulting in the poorest modeling quality. The 70\% and 80\% overlap ratios showed slight differences in values, with 70\% yielding the best results.}\label{fig13}
\end{figure}

Figure~\ref{fig13} presents the outcomes concerning the maximum and minimum values obtained from the Covariance Matrix, for each modeling, considering the different overlap sets and the distribution with 8 SBs

In the initial stages, the data acquired for 60\% overlaps were excluded due to their considerably high RMSE and Covariance Matrix values compared to the other configurations employed in this study. The elevated values can be attributed to the smaller number of images used, leading to less detail in the conducted modeling.

In configurations with overlaps of 70\% and 80\%, similarities were observed between the two analyzed quality values. There is a slight advantage for the configuration utilizing the 70\% overlap rate. The RMSE values obtained were 0.14 mm and 0.16 mm, with the Covariance Matrix values slightly favoring a 70\% overlap configuration.

Although the overlap settings of 70\% and 80\% yielded similar values for the different quality analyses of the 3D models, one element must be considered as a significant tiebreaker: the number of images used. Modeling with an 80\% overlap used 153 images, while the version with a lower overlap rate, to obtain slightly similar results, required a third fewer images (102) of the region of interest.

Configurations with constant overlapping rates of 70\% and 80\% showed satisfactory submillimetric quality levels and the choice between those setups depends on the number of images required. In projects that demand more time for the capture and modeling process, it is advantageous to choose higher overlap values. In contrast, in projects with time constraints, the configuration with 70\% overlap is adequate, despite a slight decline in quality.

Objectively speaking, the values referring to the quality of the 3D modeling, for the considered overlapping configurations, presented slight similarities, with only the quantity of images used as a possible choice factor, among the used configurations. This behavior was also observed in \cite{rupnik2015}, who observed a slight improvement in modeling to the increase in images, in contrast to the increase in processing time.

Therefore, greater overlap becomes advantageous for projects with more available time or that require greater details in the region of interest. On the other hand, in projects in which time is a dominant factor, the configuration with 70\% overlap has proved sufficient, with a slight loss of quality in environments with many details on the modeled object.

\subsection{Combined use of Vertical and Oblique Images}\label{subsec11}
Aiming to evaluate the influence of sets of vertical and oblique images in internal environments with the potential for use in structural tests, new analyses were conducted, considering combinations of vertical and oblique images.

For these evaluations, given the 1-meter capture distance, we opted to employ two different sets of oblique capture, first with the camera rotated 15° only in Yaw and, after, the camera rotated 15° only in Pitch, to the wooden board. To conduct the modeling analyses, we used clustering based on the rotation movement of the camera. The adopted models were as follows: only vertical images (V); a combination of vertical and oblique images rotated in Yaw (Y); a combination of vertical and oblique images rotated in Pitch (P); and all images (A). Table~\ref{tab5} presents the different sets of images, overlap rate, and number of images for this experiment.

\begin{table}[h]
    \caption{Settings used in this experiment, camera position, \% overlap, and number of images.}\label{tab5}
    \begin{tabular}{|c|c|c|c|c|}
    \hline
    \textbf{Vertical} & \textbf{Yaw} & \textbf{Pitch} & \% \textbf{Overlap} & \textbf{No. of Images}    \\ \hline
    X        & -   & -     & 70\%       & 102              \\ \hline
    X        & X   & -     & 70\%       & 301              \\ \hline
    X        & -   & X     & 70\%       & 272              \\ \hline
    X        & X   & X     & 70\%       & 454              \\ \hline
    X        & -   & -     & 80\%       & 153              \\ \hline
    X        & X   & -     & 80\%       & 413              \\ \hline
    X        & -   & X     & 80\%       & 391              \\ \hline
    \end{tabular}
\end{table}

\begin{figure}[h!]
    \centering
    \includegraphics[width=\textwidth]{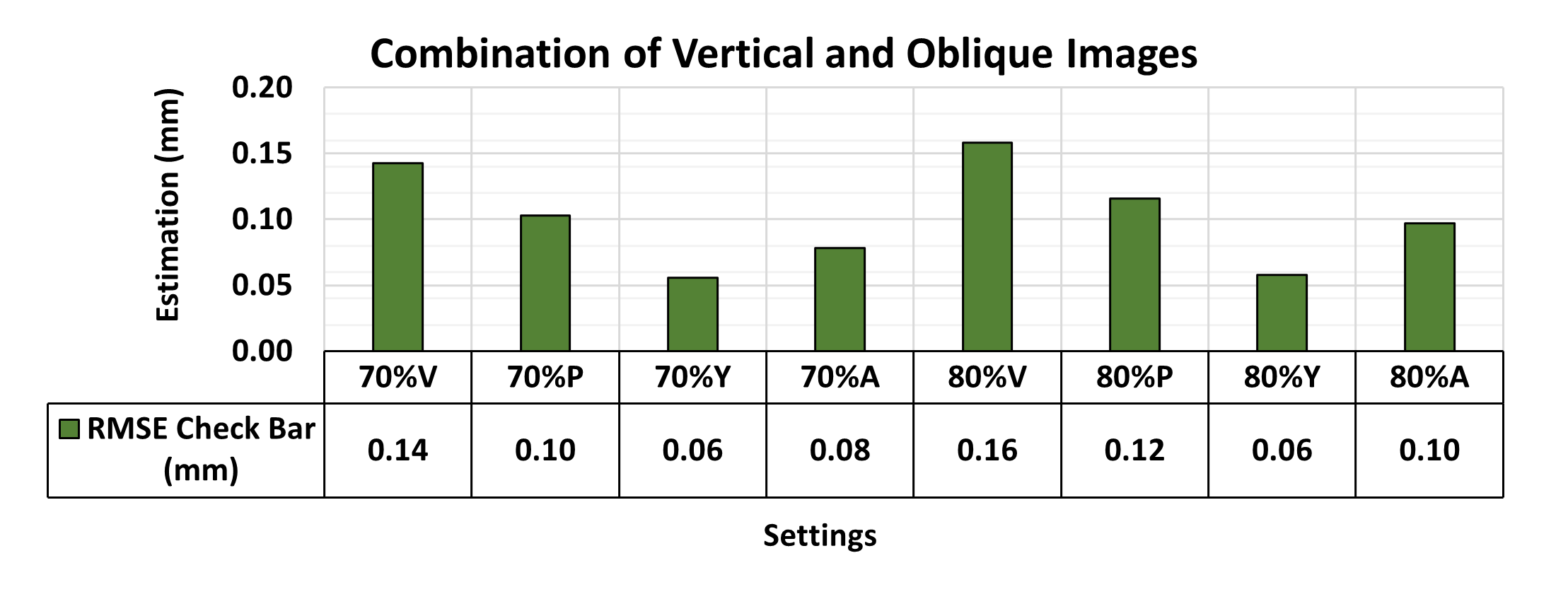}
    \caption{RMSE values of CBs for different combinations of image settings, using the maximum number of SBs and the configuration of 80\% overlap. The best results were yielded using sets of vertical images and oblique images with rotation in yaw, for both sets of overlap.}\label{fig14}
\end{figure}

Figure~\ref{fig14} presents the quality values referring to the different combinations of sets of images, for the best configurations previously determined with PSC calibration and 8 SBs and constant overlaps of 70\% and 80\%.

Based on the analysis of the RMSE values, in all the processing performed, we identified a significant improvement in the quality of the estimated 3D models, as discussed in \cite{nesbit2019} and \cite{james2014}, when using some combination of vertical images with oblique images, versus using only vertical images (70\%V and 80\%V). As observed, for both overlapping configurations, the use of sets of vertical images and oblique images with rotation in Yaw (70\%Y and 80\%Y), using 301 and 413 captures respectively, presented the best RMSE with values of 0.06 mm for both configurations of overlap.

\begin{figure}[h!]
    \centering
    \includegraphics[width=\textwidth]{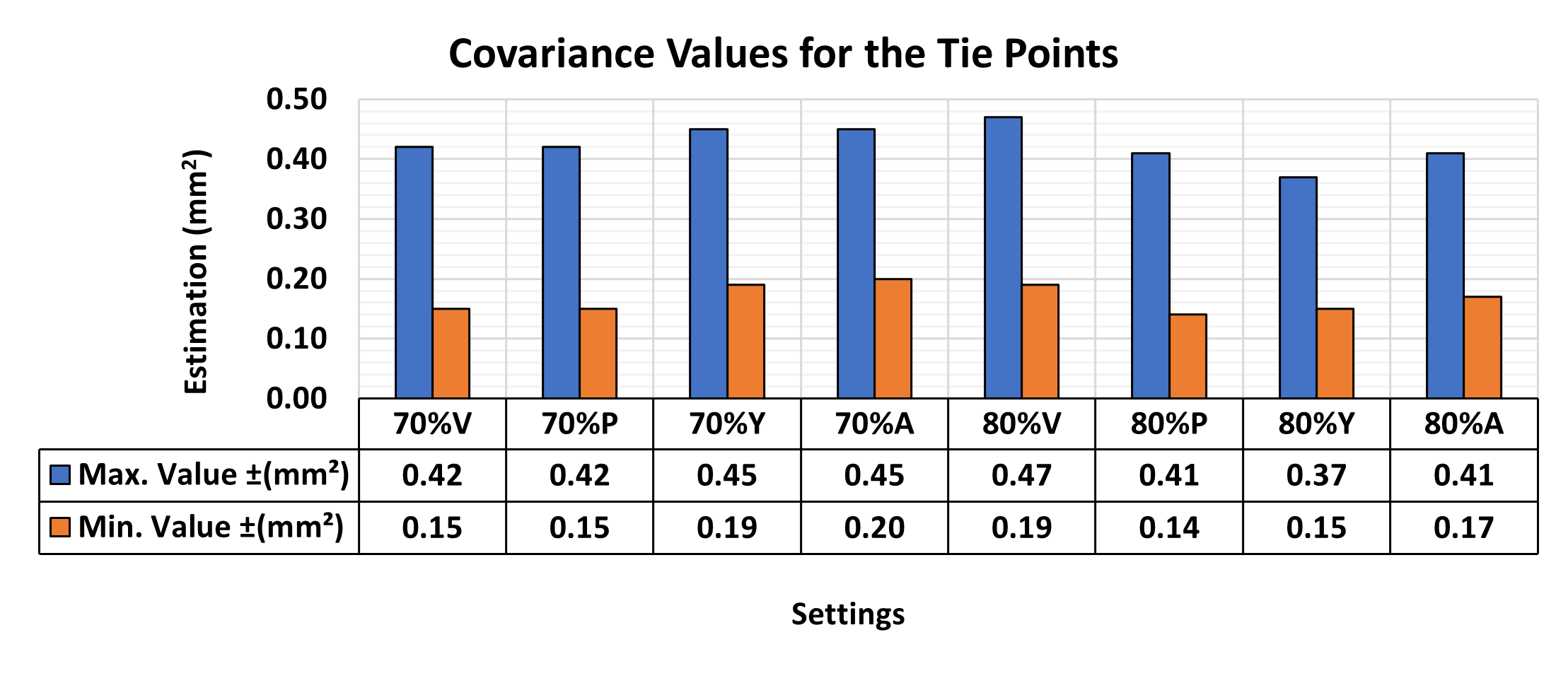}
    \caption{Maximum and minimum values obtained from the Covariance Matrix from the different sets of images. The most favorable outcomes in the experiment were observed for configuration 80\%Y, with a maximum value of 0.37 mm² and a minimum value of 0.15 mm².}\label{fig15}
\end{figure}

In terms of adjustment quality, as depicted in Figure~\ref{fig15}, the values obtained from the Covariance Matrix showed the best results in this experiment for the configuration with 80\% overlap and a combination of vertical and oblique images (80\%Y). When utilizing a 70\% overlap, only vertical images yielded the best results for maximum and minimum values, a pattern mirrored when combined with 70\%P.

The combination of vertical images and oblique images with rotation in Pitch (X-axis) for 70\%P and 80\%P models, and 272 and 391 images respectively, also presented satisfactory results, to the configuration of only vertical images. However, with a slight worsening in the RMSE values, 0.10 mm for 70\% and 0.12 mm for 80\%, to the best results. The maximum and minimum values obtained from these configurations were 0.42 mm² and 0.15 mm², for an overlap of 70\%, and 0.41 mm² and 0.14 mm² for 80\%, respectively.

Although the RMSE results showed significant improvements compared to values obtained with vertical images only, the values of the Covariance Matrix did not exhibit gains when analyzing the results of modeling with a 70\% overlap and set of oblique images. 

In the case of the 70\%P model, the result remained practically unchanged, while for the 70\%Y model, a notable deterioration in modeling quality was observed. This decline in results was reflected in the values when combining all different images (70\%A). 

However, the values obtained by this quality variable in configurations with an 80\% overlap showed improvements. For the 80\%P model, there was an insignificant improvement, but for the 80\%Y model, the improvement was on the order of 10\%, establishing itself as the most effective configuration in this analysis. Due to the quality of the 80\%P set, the combination of these two sets also achieved insignificant improvements.

The deterioration in quality assessment can be justified in configurations with a 70\% overlap concerning the board, attributed to the qualities of the images used. Specifically, this is likely due to the way the region was illuminated during the photographic capture process, resulting in variations in light across different regions. This behavior had a more pronounced impact on configurations where the overlap rate, and consequently the number of images, was lower. Consequently, as discussed earlier, a larger number of images can provide greater detail in the final modeling.

Finally, this research has demonstrated that the overlapping percentages combined with vertical and oblique images do not affect the RMSE of three-dimensional modeling. Instead, those percentages directly impacted the modeling quality. These two variables, RMSE and values maximum and minimum, obtained from the Covariance Matrix, should be considered when analyzing the quality of a 3D model, especially when aiming for submillimeter levels.

\section{Conclusion}\label{sec5}
This study analyzed the possibility of adopting the SfM computational technique, using low-cost equipment, for use in indoor environments with short capture distances and the need for submillimeter accuracy, such as that presented in laboratory structural tests.

To estimate the quality levels obtained, as well as to determine the approaches and possible improvements to be made in a real test, a set of processes was developed, referring to different photographic captures made at 1 meter from a wooden board with several SBs, to simulate a quality analysis through the RMSE values and the Covariance Matrix of products from 3D modeling.

The analyses of the quality values suggested more efficient results when using the PSC pre-calibration model, with the distribution of SBs throughout the region of interest of the modeled object. The results of the amount of overlap indicated significant advantages of the use of 70\% and 80\% overlaps for both axes when using only vertical images, with a distinct variation in the number of images for each configuration, which can influence the total time of the 3D modeling process.

The combination of vertical and oblique images, with 15° of rotation in Yaw and Pitch, to the camera position, produced superior results compared to other techniques investigated in this study. In particular, the combination of vertical and oblique images with an 80\% overlap, employing the PSC calibration model and distributing 8 SBs throughout the region of interest of the object, demonstrated significant improvements in the quality of 3D modeling.

The analyses corroborated the performance of SfM modeling for structural tests, which require the detection of submillimetric variations in their axes, with popular equipment. Research limitations, however, include the need for greater care with the lighting of the object of interest to avoid regions with light variations. Future work can involve analyses of new luminous coverings, evaluations of different materials, common in structural tests, detection of natural textures, and use of artificial ones for effective modeling.

\backmatter

\bmhead{Supplementary information}

All image datasets used in this study are available for download from the Zenodo repository via DOI: 10.5281/zenodo.11479934.

\bmhead{Acknowledgements}

The authors thank the São Carlos School of Engineering for all the support. This study was financed by the Coordenação de Aperfeiçoamento de Pessoal de Nível Superior - Brasil (CAPES) - Finance Code 001 – process number: 88882.379118/2019-01.

\section*{Declarations}

\begin{itemize}
\item Funding
This study was financed by the Coordenação de Aperfeiçoamento de Pessoal de Nível Superior - Brasil (CAPES) - Finance Code 001 – process number: 88882.379118/2019-01
\item Conflict of interest/Competing interests
Not applicable
\item Ethics approval and consent to participate
Not applicable
\item Consent for publication
Not applicable
\item Data availability 
The authors declare that the data supporting the findings of this study are available within the paper. Raw data files in alternative formats can be obtained from the corresponding author upon reasonable request. The source data for this paper are available at DOI: 10.5281/zenodo.11479934.

\item Materials availability
Not applicable
\item Code availability 
Not applicable
\item Author contribution
All authors contributed to the study, conception, and design. Francisco Roza de Moraes and Irineu da Silva performed material preparation, data collection, and analysis. Francisco Roza de Moraes wrote the first draft of the manuscript and all authors commented on previous versions of the manuscript. All authors read and approved the final manuscript.

\end{itemize}

\bibliography{sn-article}


\begin{thebibliography}{30}
\ifx \bisbn   \undefined \def \bisbn  #1{ISBN #1}\fi
\ifx \binits  \undefined \def \binits#1{#1}\fi
\ifx \bauthor  \undefined \def \bauthor#1{#1}\fi
\ifx \batitle  \undefined \def \batitle#1{#1}\fi
\ifx \bjtitle  \undefined \def \bjtitle#1{#1}\fi
\ifx \bvolume  \undefined \def \bvolume#1{\textbf{#1}}\fi
\ifx \byear  \undefined \def \byear#1{#1}\fi
\ifx \bissue  \undefined \def \bissue#1{#1}\fi
\ifx \bfpage  \undefined \def \bfpage#1{#1}\fi
\ifx \blpage  \undefined \def \blpage #1{#1}\fi
\ifx \burl  \undefined \def \burl#1{\textsf{#1}}\fi
\ifx \doiurl  \undefined \def \doiurl#1{\url{https://doi.org/#1}}\fi
\ifx \betal  \undefined \def \betal{\textit{et al.}}\fi
\ifx \binstitute  \undefined \def \binstitute#1{#1}\fi
\ifx \binstitutionaled  \undefined \def \binstitutionaled#1{#1}\fi
\ifx \bctitle  \undefined \def \bctitle#1{#1}\fi
\ifx \beditor  \undefined \def \beditor#1{#1}\fi
\ifx \bpublisher  \undefined \def \bpublisher#1{#1}\fi
\ifx \bbtitle  \undefined \def \bbtitle#1{#1}\fi
\ifx \bedition  \undefined \def \bedition#1{#1}\fi
\ifx \bseriesno  \undefined \def \bseriesno#1{#1}\fi
\ifx \blocation  \undefined \def \blocation#1{#1}\fi
\ifx \bsertitle  \undefined \def \bsertitle#1{#1}\fi
\ifx \bsnm \undefined \def \bsnm#1{#1}\fi
\ifx \bsuffix \undefined \def \bsuffix#1{#1}\fi
\ifx \bparticle \undefined \def \bparticle#1{#1}\fi
\ifx \barticle \undefined \def \barticle#1{#1}\fi
\bibcommenthead
\ifx \bconfdate \undefined \def \bconfdate #1{#1}\fi
\ifx \botherref \undefined \def \botherref #1{#1}\fi
\ifx \url \undefined \def \url#1{\textsf{#1}}\fi
\ifx \bchapter \undefined \def \bchapter#1{#1}\fi
\ifx \bbook \undefined \def \bbook#1{#1}\fi
\ifx \bcomment \undefined \def \bcomment#1{#1}\fi
\ifx \oauthor \undefined \def \oauthor#1{#1}\fi
\ifx \citeauthoryear \undefined \def \citeauthoryear#1{#1}\fi
\ifx \endbibitem  \undefined \def \endbibitem {}\fi
\ifx \bconflocation  \undefined \def \bconflocation#1{#1}\fi
\ifx \arxivurl  \undefined \def \arxivurl#1{\textsf{#1}}\fi
\csname PreBibitemsHook\endcsname

\bibitem[\protect\citeauthoryear{Anderson et~al.}{2019}]{anderson2019}
\begin{barticle}
\bauthor{\bsnm{Anderson}, \binits{K.}},
\bauthor{\bsnm{Westoby}, \binits{M.J.}},
\bauthor{\bsnm{James}, \binits{M.R.}}:
\batitle{Low-budget topographic surveying comes of age: Structure from motion photogrammetry in geography and the geosciences}.
\bjtitle{Progress in Physical Geography: Earth and Environment}
\bvolume{43}(\bissue{2}),
\bfpage{163}--\blpage{173}
(\byear{2019})
\doiurl{10.1177/0309133319837454}
\end{barticle}
\endbibitem

\bibitem[\protect\citeauthoryear{J.~J. Carrera-Hernández and Lacan}{2020}]{carrera-Hernández2020}
\begin{barticle}
\bauthor{\bsnm{J.~J.~Carrera-Hernández}, \binits{G.L.}},
\bauthor{\bsnm{Lacan}, \binits{P.}}:
\batitle{Is uav-sfm surveying ready to replace traditional surveying techniques?}
\bjtitle{International Journal of Remote Sensing}
\bvolume{41}(\bissue{12}),
\bfpage{4820}--\blpage{4837}
(\byear{2020})
\doiurl{10.1080/01431161.2020.1727049}
\end{barticle}
\endbibitem

\bibitem[\protect\citeauthoryear{Stott et~al.}{2020}]{stott2020}
\begin{barticle}
\bauthor{\bsnm{Stott}, \binits{E.}},
\bauthor{\bsnm{Williams}, \binits{R.D.}},
\bauthor{\bsnm{Hoey}, \binits{T.B.}}:
\batitle{Ground control point distribution for accurate kilometre-scale topographic mapping using an rtk-gnss unmanned aerial vehicle and sfm photogrammetry}.
\bjtitle{Drones}
\bvolume{4}(\bissue{3}),
\bfpage{55}
(\byear{2020})
\end{barticle}
\endbibitem

\bibitem[\protect\citeauthoryear{Smith and Vericat}{2015}]{smith2015}
\begin{barticle}
\bauthor{\bsnm{Smith}, \binits{M.W.}},
\bauthor{\bsnm{Vericat}, \binits{D.}}:
\batitle{From experimental plots to experimental landscapes: topography, erosion and deposition in sub-humid badlands from structure-from-motion photogrammetry}.
\bjtitle{Earth Surface Processes and Landforms}
\bvolume{40}(\bissue{12}),
\bfpage{1656}--\blpage{1671}
(\byear{2015})
\end{barticle}
\endbibitem

\bibitem[\protect\citeauthoryear{Warrick et~al.}{2017}]{warrick2017}
\begin{barticle}
\bauthor{\bsnm{Warrick}, \binits{J.A.}},
\bauthor{\bsnm{Ritchie}, \binits{A.C.}},
\bauthor{\bsnm{Adelman}, \binits{G.}},
\bauthor{\bsnm{Adelman}, \binits{K.}},
\bauthor{\bsnm{Limber}, \binits{P.W.}}:
\batitle{New techniques to measure cliff change from historical oblique aerial photographs and structure-from-motion photogrammetry}.
\bjtitle{Journal of Coastal Research}
\bvolume{33}(\bissue{1}),
\bfpage{39}--\blpage{55}
(\byear{2017})
\end{barticle}
\endbibitem

\bibitem[\protect\citeauthoryear{Zimmer et~al.}{2018}]{zimmer2018}
\begin{barticle}
\bauthor{\bsnm{Zimmer}, \binits{B.}},
\bauthor{\bsnm{Liutkus-Pierce}, \binits{C.}},
\bauthor{\bsnm{Marshall}, \binits{S.T.}},
\bauthor{\bsnm{Hatala}, \binits{K.G.}},
\bauthor{\bsnm{Metallo}, \binits{A.}},
\bauthor{\bsnm{Rossi}, \binits{V.}}:
\batitle{Using differential structure-from-motion photogrammetry to quantify erosion at the engare sero footprint site, tanzania}.
\bjtitle{Quaternary Science Reviews}
\bvolume{198},
\bfpage{226}--\blpage{241}
(\byear{2018})
\end{barticle}
\endbibitem

\bibitem[\protect\citeauthoryear{Morgan et~al.}{2017}]{morgan2017}
\begin{barticle}
\bauthor{\bsnm{Morgan}, \binits{J.A.}},
\bauthor{\bsnm{Brogan}, \binits{D.J.}},
\bauthor{\bsnm{Nelson}, \binits{P.A.}}:
\batitle{Application of structure-from-motion photogrammetry in laboratory flumes}.
\bjtitle{Geomorphology}
\bvolume{276},
\bfpage{125}--\blpage{143}
(\byear{2017})
\end{barticle}
\endbibitem

\bibitem[\protect\citeauthoryear{Scaioni et~al.}{2015}]{scaioni2015}
\begin{barticle}
\bauthor{\bsnm{Scaioni}, \binits{M.}},
\bauthor{\bsnm{Feng}, \binits{T.}},
\bauthor{\bsnm{Barazzetti}, \binits{L.}},
\bauthor{\bsnm{Previtali}, \binits{M.}},
\bauthor{\bsnm{Roncella}, \binits{R.}}:
\batitle{Image-based deformation measurement}.
\bjtitle{Applied Geomatics}
\bvolume{7},
\bfpage{75}--\blpage{90}
(\byear{2015})
\end{barticle}
\endbibitem

\bibitem[\protect\citeauthoryear{Nolan et~al.}{2015}]{nolan2015}
\begin{barticle}
\bauthor{\bsnm{Nolan}, \binits{M.}},
\bauthor{\bsnm{Larsen}, \binits{C.}},
\bauthor{\bsnm{Sturm}, \binits{M.}}:
\batitle{Mapping snow depth from manned aircraft on landscape scales at centimeter resolution using structure-from-motion photogrammetry}.
\bjtitle{The Cryosphere}
\bvolume{9}(\bissue{4}),
\bfpage{1445}--\blpage{1463}
(\byear{2015})
\end{barticle}
\endbibitem

\bibitem[\protect\citeauthoryear{Wu et~al.}{2020}]{wu2020}
\begin{barticle}
\bauthor{\bsnm{Wu}, \binits{H.}},
\bauthor{\bsnm{Zheng}, \binits{D.-f.}},
\bauthor{\bsnm{Zhang}, \binits{Y.-j.}},
\bauthor{\bsnm{Li}, \binits{D.-y.}},
\bauthor{\bsnm{Nian}, \binits{T.-k.}}:
\batitle{A photogrammetric method for laboratory-scale investigation on 3d landslide dam topography}.
\bjtitle{Bulletin of Engineering Geology and the Environment}
\bvolume{79},
\bfpage{4717}--\blpage{4732}
(\byear{2020})
\end{barticle}
\endbibitem

\bibitem[\protect\citeauthoryear{Cucchiaro et~al.}{2020}]{cucchiaro2020}
\begin{bchapter}
\bauthor{\bsnm{Cucchiaro}, \binits{S.}},
\bauthor{\bsnm{Fallu}, \binits{D.J.}},
\bauthor{\bsnm{Zhao}, \binits{P.}},
\bauthor{\bsnm{Waddington}, \binits{C.}},
\bauthor{\bsnm{Cockcroft}, \binits{D.}},
\bauthor{\bsnm{Tarolli}, \binits{P.}},
\bauthor{\bsnm{Brown}, \binits{A.G.}}:
\bctitle{Sfm photogrammetry for geoarchaeology}.
In: \bbtitle{Developments in Earth Surface Processes}
vol. \bseriesno{23},
pp. \bfpage{183}--\blpage{205}.
\bpublisher{Elsevier}, \blocation{???}
(\byear{2020})
\end{bchapter}
\endbibitem

\bibitem[\protect\citeauthoryear{Moyano et~al.}{2020}]{moyano2020}
\begin{barticle}
\bauthor{\bsnm{Moyano}, \binits{J.}},
\bauthor{\bsnm{Nieto-Juli{\'a}n}, \binits{J.E.}},
\bauthor{\bsnm{Bienvenido-Huertas}, \binits{D.}},
\bauthor{\bsnm{Mar{\'\i}n-Garc{\'\i}a}, \binits{D.}}:
\batitle{Validation of close-range photogrammetry for architectural and archaeological heritage: Analysis of point density and 3d mesh geometry}.
\bjtitle{Remote sensing}
\bvolume{12}(\bissue{21}),
\bfpage{3571}
(\byear{2020})
\end{barticle}
\endbibitem

\bibitem[\protect\citeauthoryear{Nesbit and Hugenholtz}{2019}]{nesbit2019}
\begin{barticle}
\bauthor{\bsnm{Nesbit}, \binits{P.R.}},
\bauthor{\bsnm{Hugenholtz}, \binits{C.H.}}:
\batitle{Enhancing uav--sfm 3d model accuracy in high-relief landscapes by incorporating oblique images}.
\bjtitle{Remote Sensing}
\bvolume{11}(\bissue{3}),
\bfpage{239}
(\byear{2019})
\end{barticle}
\endbibitem

\bibitem[\protect\citeauthoryear{Verma and Bourke}{2019}]{verma2019}
\begin{barticle}
\bauthor{\bsnm{Verma}, \binits{A.K.}},
\bauthor{\bsnm{Bourke}, \binits{M.C.}}:
\batitle{A method based on structure-from-motion photogrammetry to generate sub-millimetre-resolution digital elevation models for investigating rock breakdown features}.
\bjtitle{Earth Surface Dynamics}
\bvolume{7}(\bissue{1}),
\bfpage{45}--\blpage{66}
(\byear{2019})
\end{barticle}
\endbibitem

\bibitem[\protect\citeauthoryear{Pe{\~n}a-Villasen{\'\i}n et~al.}{2019}]{pena2019}
\begin{barticle}
\bauthor{\bsnm{Pe{\~n}a-Villasen{\'\i}n}, \binits{S.}},
\bauthor{\bsnm{Gil-Docampo}, \binits{M.}},
\bauthor{\bsnm{Ortiz-Sanz}, \binits{J.}}:
\batitle{Professional sfm and tls vs a simple sfm photogrammetry for 3d modelling of rock art and radiance scaling shading in engraving detection}.
\bjtitle{Journal of Cultural Heritage}
\bvolume{37},
\bfpage{238}--\blpage{246}
(\byear{2019})
\end{barticle}
\endbibitem

\bibitem[\protect\citeauthoryear{Carrivick et~al.}{2016}]{carrivick2016}
\begin{bbook}
\bauthor{\bsnm{Carrivick}, \binits{J.L.}},
\bauthor{\bsnm{Smith}, \binits{M.W.}},
\bauthor{\bsnm{Quincey}, \binits{D.J.}}:
\bbtitle{Structure from Motion in the Geosciences}.
\bpublisher{John Wiley \& Sons}, \blocation{???}
(\byear{2016})
\end{bbook}
\endbibitem

\bibitem[\protect\citeauthoryear{Westoby et~al.}{2012}]{westoby2012}
\begin{barticle}
\bauthor{\bsnm{Westoby}, \binits{M.J.}},
\bauthor{\bsnm{Brasington}, \binits{J.}},
\bauthor{\bsnm{Glasser}, \binits{N.F.}},
\bauthor{\bsnm{Hambrey}, \binits{M.J.}},
\bauthor{\bsnm{Reynolds}, \binits{J.M.}}:
\batitle{‘structure-from-motion’photogrammetry: A low-cost, effective tool for geoscience applications}.
\bjtitle{Geomorphology}
\bvolume{179},
\bfpage{300}--\blpage{314}
(\byear{2012})
\end{barticle}
\endbibitem

\bibitem[\protect\citeauthoryear{Gerke et~al.}{2016}]{gerke2016}
\begin{barticle}
\bauthor{\bsnm{Gerke}, \binits{M.}},
\bauthor{\bsnm{Nex}, \binits{F.}},
\bauthor{\bsnm{Remondino}, \binits{F.}},
\bauthor{\bsnm{Jacobsen}, \binits{K.}},
\bauthor{\bsnm{Kremer}, \binits{J.}},
\bauthor{\bsnm{Karel}, \binits{W.}},
\bauthor{\bsnm{Huf}, \binits{H.}},
\bauthor{\bsnm{Ostrowski}, \binits{W.}}:
\batitle{Orientation of oblique airborne image sets-experiences from the isprs/eurosdr benchmark on multi-platform photogrammetry}.
\bjtitle{The International Archives of the Photogrammetry, Remote Sensing and Spatial Information Sciences 41-B1}
\bvolume{41},
\bfpage{185}--\blpage{191}
(\byear{2016})
\end{barticle}
\endbibitem

\bibitem[\protect\citeauthoryear{James and Robson}{2014}]{james2014}
\begin{barticle}
\bauthor{\bsnm{James}, \binits{M.R.}},
\bauthor{\bsnm{Robson}, \binits{S.}}:
\batitle{Mitigating systematic error in topographic models derived from uav and ground-based image networks}.
\bjtitle{Earth Surface Processes and Landforms}
\bvolume{39}(\bissue{10}),
\bfpage{1413}--\blpage{1420}
(\byear{2014})
\end{barticle}
\endbibitem

\bibitem[\protect\citeauthoryear{Rupnik et~al.}{2015}]{rupnik2015}
\begin{barticle}
\bauthor{\bsnm{Rupnik}, \binits{E.}},
\bauthor{\bsnm{Nex}, \binits{F.}},
\bauthor{\bsnm{Toschi}, \binits{I.}},
\bauthor{\bsnm{Remondino}, \binits{F.}}:
\batitle{Aerial multi-camera systems: Accuracy and block triangulation issues}.
\bjtitle{ISPRS Journal of Photogrammetry and Remote Sensing}
\bvolume{101},
\bfpage{233}--\blpage{246}
(\byear{2015})
\end{barticle}
\endbibitem

\bibitem[\protect\citeauthoryear{Ridolfi et~al.}{2017}]{ridolfi2017}
\begin{barticle}
\bauthor{\bsnm{Ridolfi}, \binits{E.}},
\bauthor{\bsnm{Buffi}, \binits{G.}},
\bauthor{\bsnm{Venturi}, \binits{S.}},
\bauthor{\bsnm{Manciola}, \binits{P.}}:
\batitle{Accuracy analysis of a dam model from drone surveys}.
\bjtitle{Sensors}
\bvolume{17}(\bissue{8}),
\bfpage{1777}
(\byear{2017})
\end{barticle}
\endbibitem

\bibitem[\protect\citeauthoryear{Garcia and Oliveira}{2021}]{garcia2021}
\begin{barticle}
\bauthor{\bsnm{Garcia}, \binits{M.V.Y.}},
\bauthor{\bsnm{Oliveira}, \binits{H.C.d.}}:
\batitle{The influence of flight configuration, camera calibration, and ground control points for digital terrain model and orthomosaic generation using unmanned aerial vehicles imagery}.
\bjtitle{Boletim de Ci{\^e}ncias Geod{\'e}sicas}
\bvolume{27},
\bfpage{2021015}
(\byear{2021})
\end{barticle}
\endbibitem

\bibitem[\protect\citeauthoryear{Segantine and Silva}{2015}]{segantine2015}
\begin{bbook}
\bauthor{\bsnm{Segantine}, \binits{P.}},
\bauthor{\bsnm{Silva}, \binits{I.}}:
\bbtitle{Topografia Para Engenharia: Teoria e Pr{\'a}tica de Geom{\'a}tica}
vol. \bseriesno{1}.
\bpublisher{Elsevier Brasil},
\blocation{São Paulo}
(\byear{2015})
\end{bbook}
\endbibitem

\bibitem[\protect\citeauthoryear{Leon et~al.}{2015}]{leon2015}
\begin{barticle}
\bauthor{\bsnm{Leon}, \binits{J.X.}},
\bauthor{\bsnm{Roelfsema}, \binits{C.M.}},
\bauthor{\bsnm{Saunders}, \binits{M.I.}},
\bauthor{\bsnm{Phinn}, \binits{S.R.}}:
\batitle{Measuring coral reef terrain roughness using ‘structure-from-motion’close-range photogrammetry}.
\bjtitle{Geomorphology}
\bvolume{242},
\bfpage{21}--\blpage{28}
(\byear{2015})
\end{barticle}
\endbibitem

\bibitem[\protect\citeauthoryear{James et~al.}{2017}]{james20173}
\begin{barticle}
\bauthor{\bsnm{James}, \binits{M.R.}},
\bauthor{\bsnm{Robson}, \binits{S.}},
\bauthor{\bsnm{Smith}, \binits{M.W.}}:
\batitle{3-d uncertainty-based topographic change detection with structure-from-motion photogrammetry: precision maps for ground control and directly georeferenced surveys}.
\bjtitle{Earth Surface Processes and Landforms}
\bvolume{42}(\bissue{12}),
\bfpage{1769}--\blpage{1788}
(\byear{2017})
\end{barticle}
\endbibitem

\bibitem[\protect\citeauthoryear{Tinkham and Swayze}{2021}]{tinkham2021}
\begin{barticle}
\bauthor{\bsnm{Tinkham}, \binits{W.T.}},
\bauthor{\bsnm{Swayze}, \binits{N.C.}}:
\batitle{Influence of agisoft metashape parameters on uas structure from motion individual tree detection from canopy height models}.
\bjtitle{Forests}
\bvolume{12}(\bissue{2}),
\bfpage{250}
(\byear{2021})
\end{barticle}
\endbibitem

\bibitem[\protect\citeauthoryear{{Starret Company}}{2007}]{Starret}
\begin{bbook}
\bauthor{\bsnm{{Starret Company}}}:
\bbtitle{EC799 Electronic Calipers 165}.
\bpublisher{Starret L.S.},
\blocation{Athol, Massachusetts}
(\byear{2007}).
\bcomment{Starret L.S.}
\end{bbook}
\endbibitem

\bibitem[\protect\citeauthoryear{Luhmann et~al.}{2023}]{luhmann2023}
\begin{bbook}
\bauthor{\bsnm{Luhmann}, \binits{T.}},
\bauthor{\bsnm{Robson}, \binits{S.}},
\bauthor{\bsnm{Kyle}, \binits{S.}},
\bauthor{\bsnm{Boehm}, \binits{J.}}:
\bbtitle{Close-range Photogrammetry and 3D Imaging}.
\bpublisher{Walter de Gruyter GmbH \& Co KG},
\blocation{Berlin, Boston}
(\byear{2023})
\end{bbook}
\endbibitem

\bibitem[\protect\citeauthoryear{Hartley and Zisserman}{2003}]{hartley2003}
\begin{bbook}
\bauthor{\bsnm{Hartley}, \binits{R.}},
\bauthor{\bsnm{Zisserman}, \binits{A.}}:
\bbtitle{Multiple View Geometry in Computer Vision}.
\bpublisher{Cambridge university press},
\blocation{New York}
(\byear{2003})
\end{bbook}
\endbibitem

\bibitem[\protect\citeauthoryear{{ Agisoft LLC }}{2022}]{agisoft185}
\begin{botherref}
\oauthor{\bsnm{{ Agisoft LLC }}}:
Metashape Professional Edition.
\url{https://www.agisoft.com/}
\end{botherref}
\endbibitem

\end{thebibliography}

\end{document}